\documentclass[journal]{IEEEtran}
\usepackage[switch]{lineno}
\usepackage{ifpdf}

\usepackage{cite}

\ifCLASSINFOpdf
      \usepackage[pdftex]{graphicx}
\else
  \usepackage[dvips]{graphicx}
\fi
\usepackage{overpic}

\usepackage[cmex10]{amsmath}
\usepackage{algorithmic}
\usepackage{algorithm}
\usepackage{graphicx}
\usepackage{stfloats}

\usepackage{array}

\ifCLASSOPTIONcompsoc
     \usepackage[caption=false,font=normalsize,labelfont=sf,textfont=sf]{subfig}
\else
     \usepackage[caption=false,font=footnotesize]{subfig}
\fi
\usepackage{fixltx2e}

\usepackage{stfloats}

\ifCLASSOPTIONcaptionsoff
  \usepackage[nomarkers]{endfloat}
 \let\MYoriglatexcaption\caption
 \renewcommand{\caption}[2][\relax]{\MYoriglatexcaption[#2]{#2}}
\fi

\usepackage{url}

\usepackage{color}

\begin{document}
%
\title{PISA: Pixelwise Image Saliency by Aggregating Complementary Appearance Contrast Measures with Edge-Preserving Coherence}
%
%
%

\author{Keze Wang, ~Liang Lin, ~Jiangbo Lu, ~Chenglong Li, ~Keyang Shi\thanks{This work was supported in part by Guangdong Natural Science Foundation under Grant S2013050014548 and Grant 2014A030313201, in part by Special Project on Integration of Industry, Education and Research of Guangdong Province under Grant. 2012B091000101, in part by Program of Guangzhou Zhujiang Star of Science and Technology under Grant 2013J2200067. This work is also supported by the research grant for the Human-Centered Cyber-physical Systems Programme at the Advanced Digital Sciences Center from Singapore's Agency for Science, Technology and Research (A*STAR). The corresponding author is L. Lin. } 
\IEEEcompsocitemizethanks{ \IEEEcompsocthanksitem K. Wang, L. Lin, and K. Shi are with the Sun Yat-sen University, Guangzhou, China. Email: wangkeze@alumni.sysu.edu.cn, linliang@ieee.org. 
\IEEEcompsocthanksitem J. Lu is with the Advanced Digital Sciences Center, Singapore. Email: jiangbo.lu@adsc.com.sg. 
\IEEEcompsocthanksitem C. Li is with the Anhui University, Hefei, China. Email: lcl1314@foxmail.com.}
}


\markboth{IEEE TRANSACTIONS ON IMAGE PROCESSING, 2014.}%
{K. Wang \MakeLowercase{\textit{et al.}}: PISA: Pixelwise Image Saliency by Aggregating Complementary Appearance Contrast Measures with Edge-Preserving Coherence}

\maketitle

\begin{abstract}
Driven by recent vision and graphics applications such as image segmentation and object recognition, computing pixel-accurate saliency values to uniformly highlight foreground objects becomes increasingly important. In this paper, we propose a unified framework called PISA, which stands for Pixelwise Image Saliency Aggregating various bottom-up cues and priors. It generates spatially coherent yet detail-preserving, pixel-accurate and fine-grained saliency, and overcomes the limitations of previous methods which use homogeneous superpixel-based and color only treatment. PISA aggregates multiple saliency cues in a global context such as complementary color and structure contrast measures with their spatial priors in the image domain. The saliency confidence is further jointly modeled with a neighborhood consistence constraint into an energy minimization formulation, in which each pixel will be evaluated with multiple hypothetical saliency levels. Instead of using global discrete optimization methods, we employ the cost-volume filtering technique to solve our formulation, assigning the saliency levels smoothly while preserving the edge-aware structure details. In addition, a faster version of PISA is developed using a gradient-driven image sub-sampling strategy to greatly improve the runtime efficiency while keeping comparable detection accuracy. Extensive experiments on a number of public datasets suggest that PISA convincingly outperforms other state-of-the-art approaches. In addition, with this work we also create a new dataset containing $800$ commodity images for evaluating saliency detection. 
\end{abstract}


%
\IEEEpeerreviewmaketitle

\section{Introduction}
\label{sec:intro}
\IEEEPARstart{S}{aliency} detection aims at highlighting salient foreground objects automatically from the background, and has received increasing attentions for many computer vision and graphics applications such as object recognition~\cite{salobj-iccv11}, content-aware image retargeting~\cite{ImgTar1-tog08}, video compression~\cite{videoCom-tip14} and image classification~\cite{ImgCla-tip14}. Driven by these recent applications, saliency detection has also evolved to aim at assigning pixel-accurate saliency values, going far beyond its early goal of mimicing human eye fixation. Due to lacking of a rigorous definition of saliency itself, inferring the (pixel-accurate) saliency assignment for diversified natural images without any user intervention is a highly ill-posed problem. To tackle this problem, a myriad of computational models~\cite{it-pami98,ca-cvpr10,gs-eccv12, GOVS-iccv11, l2d-pami11, KDE-icip2010, DRFI-cvpr13, blm-tip13, si-tip14, ST-tip14} have been proposed using various principles or priors ranging from high-level biological vision~\cite{hm-hn1985} to low-level image properties~\cite{sr-cvpr07}. Focusing on bottom-up, low-level saliency computation models in this paper, we identify several remaining issues to be addressed though existing models have demonstrated impressive results.

\begin{figure*}
\vspace{-25pt}
	\hspace{-6pt}
   \centering
			\subfloat[Original]	
			{ 					
				{\label{fig:original1}
				}
				\includegraphics[width = 0.243 \columnwidth]{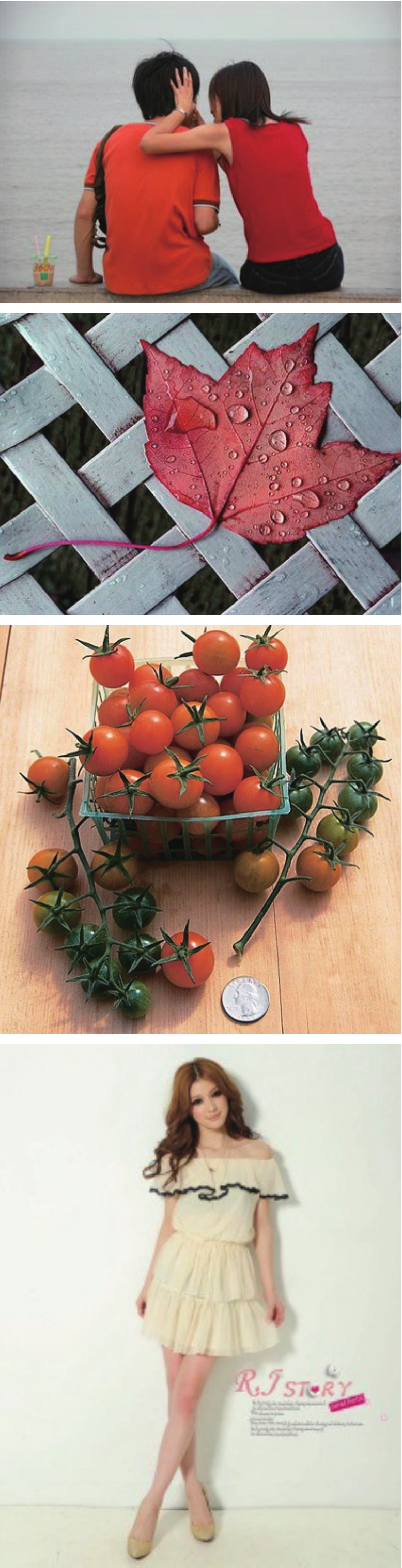}
			}	
			\hspace{-8.5pt}
			\subfloat[Color info.]	
			{ 	
				\label{fig:color1}
				\includegraphics[width = 0.243 \columnwidth]{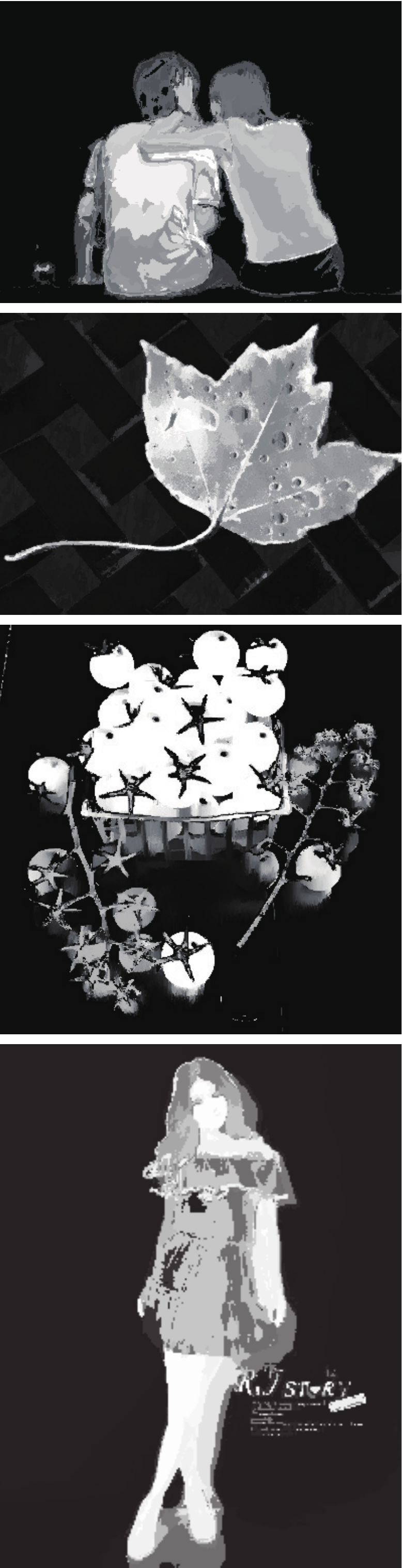}
			}
			\hspace{-8.5pt}
			\subfloat[Structure info.]	
			{ 	
				\label{fig:gradient1}
				\includegraphics[width = 0.243 \columnwidth]{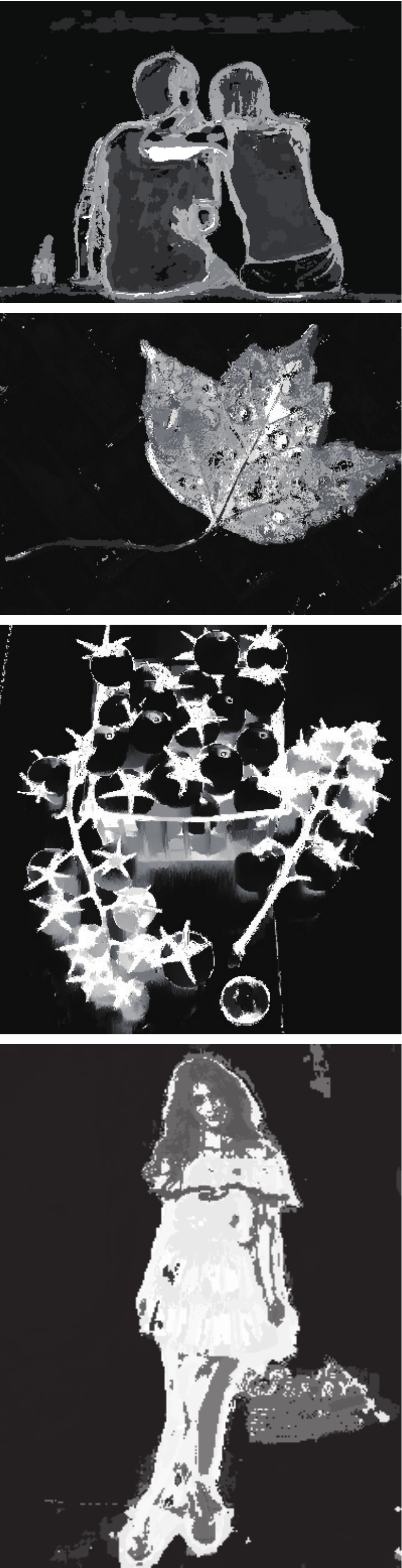}
			}
			\hspace{-8.5pt}
			\subfloat[PISA]	
			{ 	
				\label{fig:pisa1}
				\includegraphics[width = 0.243 \columnwidth]{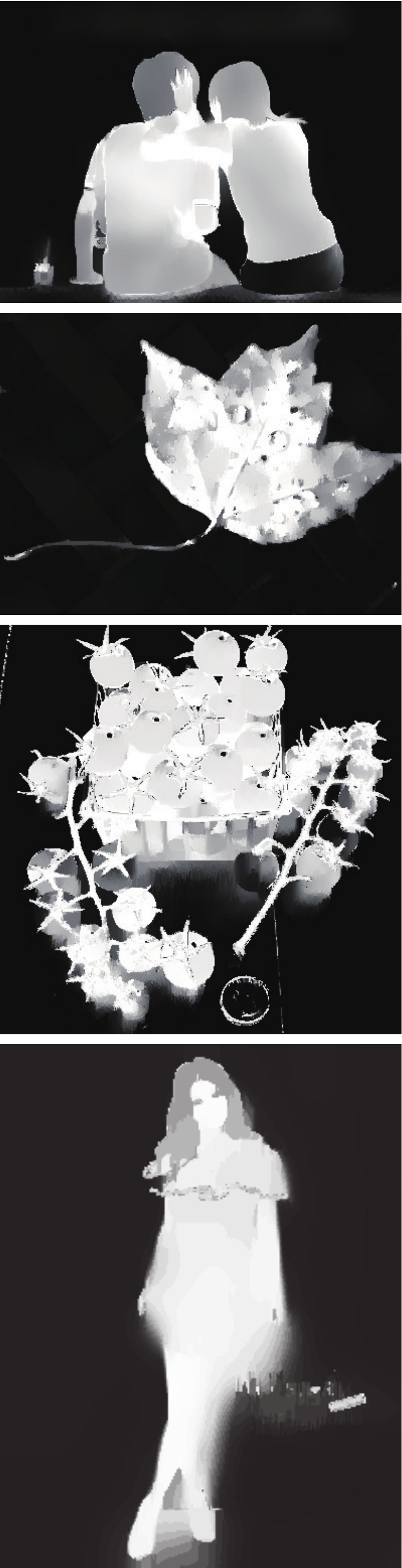}
				\includegraphics[width = 0.00766 \columnwidth]{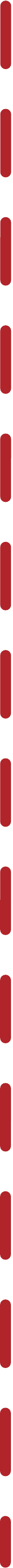}
			}
			\hspace{-6.5pt}
			\subfloat[CA\cite{ca-cvpr10}]	
			{ 	
				\label{fig:ca1}
				\includegraphics[width = 0.243 \columnwidth]{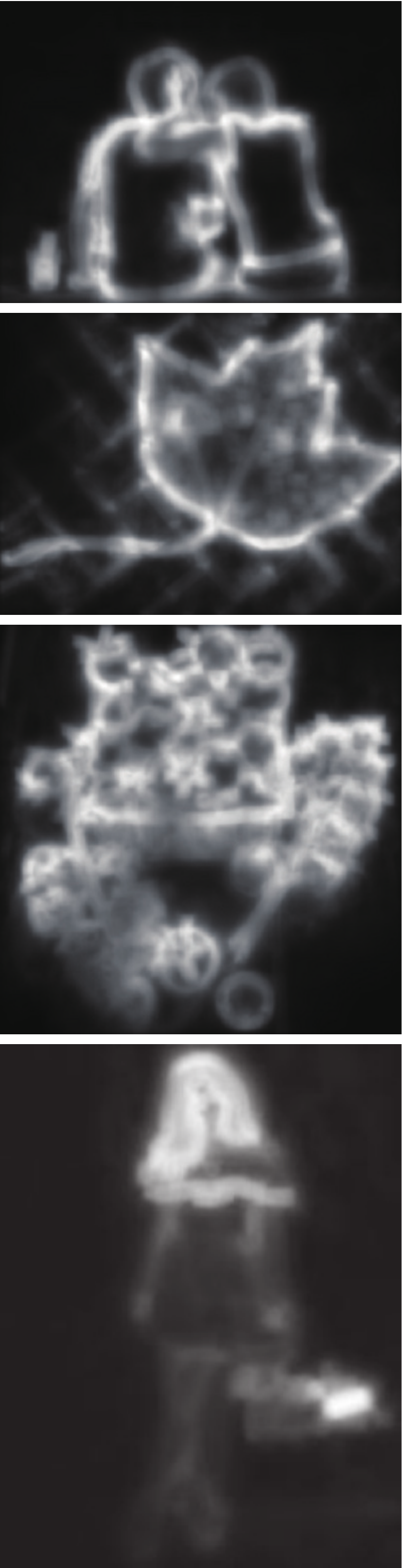}
			}
			\hspace{-8.5pt}
			\subfloat[HC\cite{hcrc-cvpr11}]	
			{ 	
				\label{fig:hc1}
				\includegraphics[width = 0.243 \columnwidth]{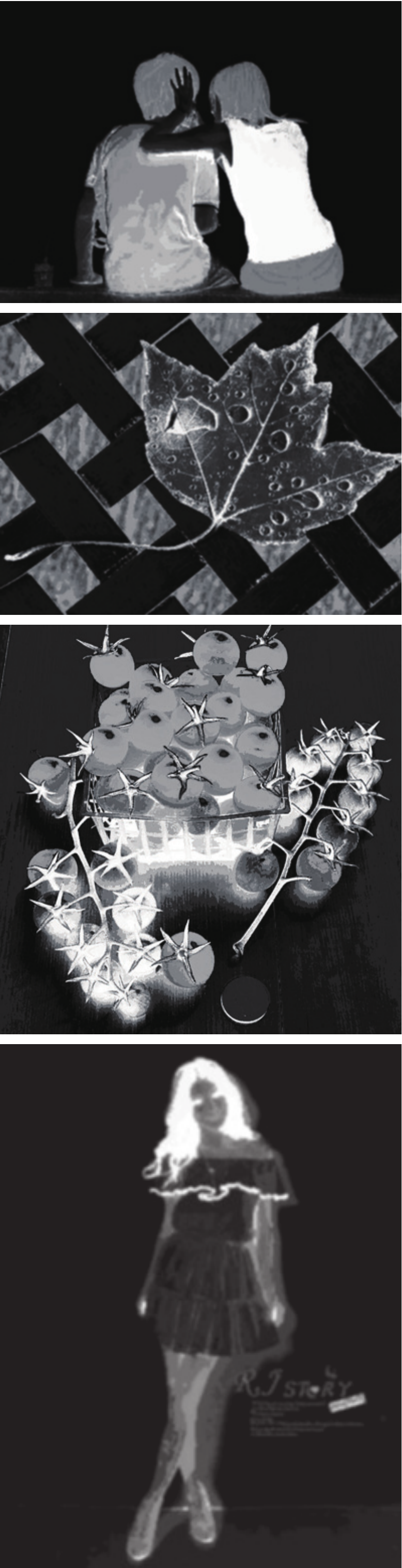}
			}
			\hspace{-8.5pt}
			\subfloat[RC\cite{hcrc-cvpr11}]	
			{ 	
				\label{fig:rc1}
				\includegraphics[width = 0.243 \columnwidth]{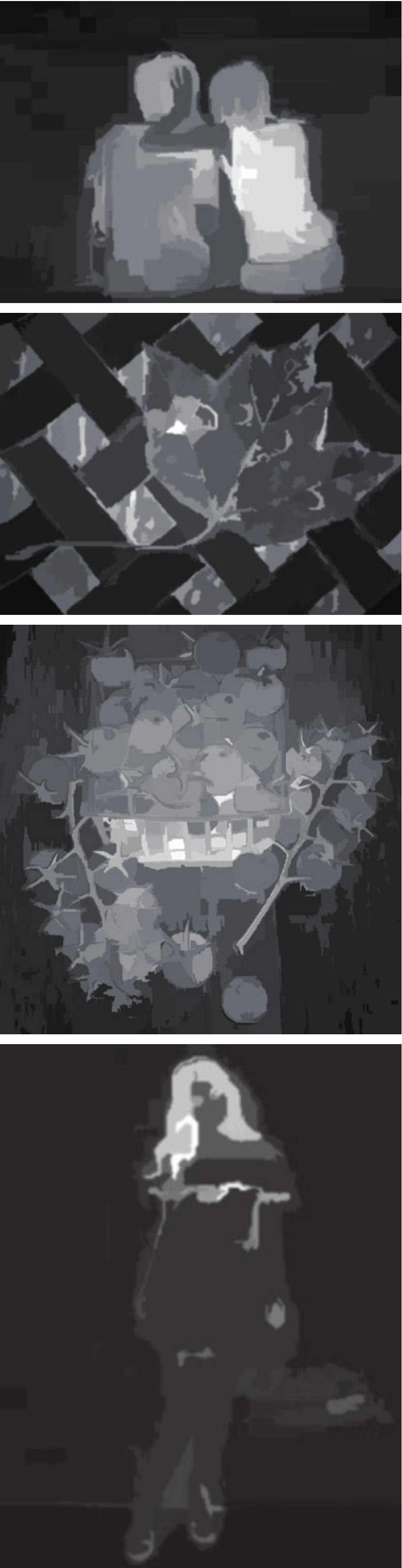}
			}
			\hspace{-8.5pt}
			\subfloat[SF\cite{sf-cvpr12}]	
			{ 	
				\label{fig:sf1}
				\includegraphics[width = 0.243 \columnwidth]{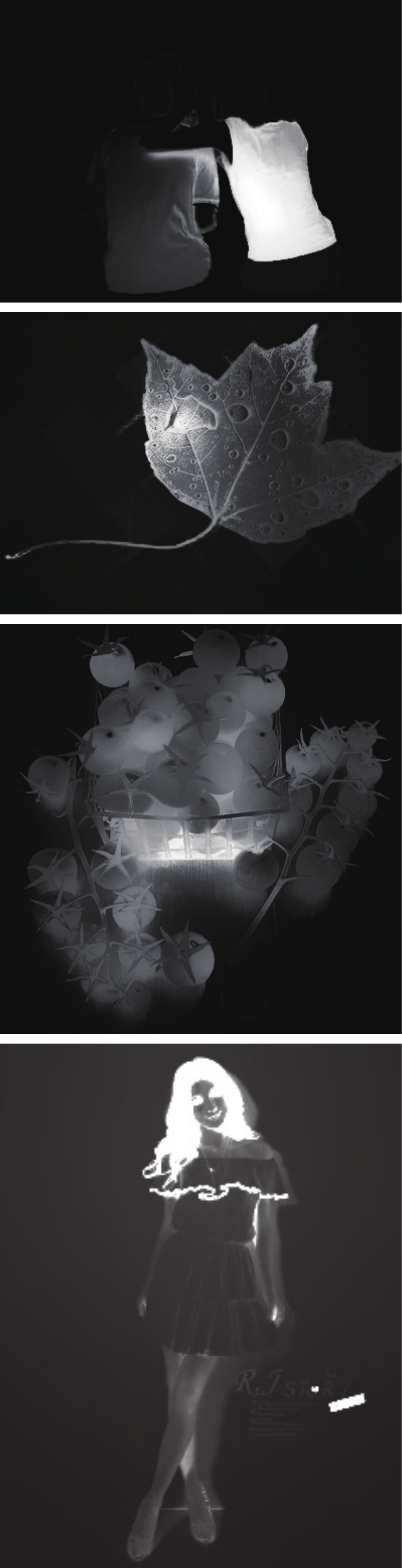}
			}
   \caption
   {Saliency maps computed on (a) four example images by (d) the proposed PISA method and (e-h) a few competing bottom-up saliency detection methods~\cite{ca-cvpr10, hcrc-cvpr11, sf-cvpr12}. The results generated by PISA with only color/structure contrast feature are shown in (b/c).}\label{fig:fig1}
   \vspace{-15pt}
   \end{figure*}

\begin{itemize}
{\item How to uniformly highlight the salient objects.}
Natural images usually contain diverse patterns (i.e. rich appearances) so that the saliency computed through the bottom-up feature extraction could be discrete or incomplete without regard to salient objects. Like other low-level vision tasks (e.g., image segmentation), most existing saliency models were built upon color information only, and they may degenerate when similar colors distribute on both foreground and background objects, e.g., Fig.~\ref{fig:fig1}(fourth row: f-h). Moreover, these approaches~\cite{hcrc-cvpr11,ca-cvpr10,sf-cvpr12} may render some elements inside a salient object as non-salient or some elements of the background as salient, due to their shortcoming on handling inhomogeneous structures in foreground (e.g., Fig.~\ref{fig:fig1}(third row: e-h)) and background (e.g., Fig.~\ref{fig:fig1}(second row: e-h)).

{\item How to make the saliency values coherent with image content.}
Several saliency detection approaches demonstrated impressive results on generating pixelwise saliency maps~\cite{hcrc-cvpr11,ca-cvpr10,sf-cvpr12}. They usually assign the saliency values based on the over-segmentation of images (i.e. small regions or superpixels), and further exploit the post-relaxation (e.g. local filtering) to smooth the saliency values over pixels. However, the image segmentation may introduce errors in processing complex image content (e.g., local cluttered textures), upon which the incompatibility with saliency values and object details could be caused by the post-relaxation step. These phenomenons are exhibited with the examples in Fig.~\ref{fig:fig1}(first row: g-h).
\end{itemize}

Inspired by the insights and lessons from a significant amount of previous work as well as several priors supported by psychological evidences and observations of natural images, we address these above mentioned challenges in a more holistic manner. In particular, we propose a unified framework called PISA, which stands for Pixelwise Image Saliency Aggregating complementary saliency cues. It enables to generate spatially coherent yet detail-preserving, pixel-accurate and fine-grained image saliency. In the following, we briefly discuss the motivations and main components of PISA.

{\it i) Complementary appearance features for measuring saliency.} Though color information is a popular saliency cue used dominantly in many methods~\cite{hcrc-cvpr11, sf-cvpr12, RegionSaliency, gc-iccv13}, other influential factors do exist, which can also be used to make salient pixels or regions outstanding, even these pixels or regions are not unique or rare by color information. For instance, they can have unique appearance features in edge/texture patterns~\cite{it-pami98}, demonstrating distinct contrast expressed by structure information. In fact, color and structure can be complementary to each other to provide more informative evidences for extracting complete salient objects. In addition, it is known from the perceptual research~\cite{RegAtt-JoN03} that different local receptive fields are associated with different kinds of visual stimuli, so local analysis regions where saliency cues are extracted should be adapted to match specific image attributes.

Instead of using color only treatment, PISA directly performs saliency modeling for each individual pixel on two complementary cues (i.e. color and structure features) and makes use of densely overlapping, feature-adaptive observations for saliency confidence computation. Fig.~\ref{fig:fig1} shows a few motivating examples that highlight the advantage of our PISA method, compared with some leading methods~\cite{hcrc-cvpr11,sf-cvpr12,ca-cvpr10}.

{\it ii) Non-parametric feature modeling in a global context.} Existing saliency detection approaches usually group image pixels based on local small regions or superpixels~\cite{hcrc-cvpr11, sf-cvpr12, blm-tip13}, which could give rise to less informative saliency measures. In contrast, using non-local approaches to summarize the extracted features~\cite{luoping-pr12, ruimao-tcsvt15,xiaolong-pami14} tends to be more robust and reasonable than those of local homogeneous superpixel-based methods, and its advantage has been demonstrated in recent works~\cite{pisa-cvpr13,gc-iccv13}.

Rather than using superpixel-based representations, we propose to compute the saliency confidence by considering both the global appearance contrast in the feature space as well as the image domain smoothness. Specifically, we first group all image pixels by summarizing their extracted features (i.e. either the color or structure histograms), and model the saliency confidence according to the global rarity (i.e. uniqueness) of the pixel group in the color/structure feature space. Meanwhile, we further impose the spatial priors, including the center preference and boundary exclusion in the image domain to complete the saliency modeling for each pixel.

{\it iii) Fine-grained saliency assignment.} Many high level tasks prefer generating more abundant and fine-grained saliency maps (i.e. each pixel can be assigned with several saliency levels). Pixel-accurate saliency maps are often required to be spatially coherent with discontinuities well aligned to image edges, according to existing studies~\cite{sf-cvpr12,acc-pr13}. In particular, the spatial connectivity and correlation involved in neighborhood pixels should be preserved in saliency computing.

In this work, we pose the fine-grained saliency assignment as a multiple labeling problem, in which the appearance contrast based saliency measure is jointly modeled with the neighborhood coherence constraint. The resulting target function can be minimized by using global discrete labeling optimizers such as graph cuts~\cite{graphcut-pami04} or belief propagation~\cite{bp-cvpr04}. These methods, however, are often relatively time-consuming and do not scale well to fine-grained labeling (i.e. a large space of labels). Some other continuous approaches are efficient but usually require a restricted form of the energy function. In this paper, we employ a recently proposed filter-based method, namely cost-volume filtering~\cite{fcv-pami13}, to smoothly assign the saliency levels while preserving structural coherence (i.e. keeping the edges and boundaries of salient objects).

To balance the accuracy-efficiency trade-off, we also propose a faster version called F-PISA. It first performs saliency computation for a feature-driven, subsampled image grid, and then uses an adaptive upsampling scheme with the color image as the guidance signal to recover a full-resolution saliency map. Compared to segmentation-based saliency methods~\cite{sf-cvpr12}, our F-PISA method reduces the computational complexity similarly by considering a coarse image grid, while having the advantage of utilizing image structural information for saliency reasoning over~\cite{sf-cvpr12}. Our extensive experiments on six public benchmarks demonstrate the superior detection accuracy and competitive runtime speed of our approaches over the state-of-the-arts. Moreover, we construct a new and meaningful database of image saliency including real commodity images from online shops\footnote{The dataset and source code of PISA can be downloaded at\\ \url{http://vision.sysu.edu.cn/project/PISA/}}.

The remainder of the paper is organized as follows: Sect.~\ref{sec:related} reviews related works of saliency detection. Sect.~\ref{sec:formul} introduces the proposed framework and its main components. More details for inference and implementation are discussed in Sect.~\ref{sec:alg}. Extensive experimental evaluations and comparisons are presented in Sect.~\ref{sec:exper}. The paper concludes in Sect.~\ref{sec:conclusion}.

\section{Related Work}
\label{sec:related}

Recently, numerous bottom-up saliency detection models have been proposed for explaining visual attention based on different mathematical principles or priors. We classify most of the previous methods into two basic classes depending on the way that saliency cues are defined: {\it contrast priors} and {\it background priors}~\cite{gs-eccv12}. Assuming that saliency is unique and rare in appearance, contrast priors have been widely adopted in many previous methods to model the appearance contrast between foreground salient objects and the background. Itti {\em et al.}~\cite{it-pami98} presented a bottom-up method in which an input image is represented with three features including color, intensity and orientation in different scales. Achanta {\em et al.}~\cite{ft-cvpr09} proposed a frequency-tuned method that defines the saliency likelihood of each pixel based on its difference from the average image color by exploiting the center-prior principle. Goferman {\em et al.}~\cite{ca-cvpr10} used a patch based approach to incorporate global properties to highlight salient objects along with their contexts. However, due to using the local contrast only, it tends to produce higher salient values near edges. To highlight the entire object, Cheng {\em et al.}~\cite{hcrc-cvpr11} presented color histogram contrast (HC) in the $Lab$ color space and region contrast (RC) in a global scope. Perazzi {\em et al.}~\cite{sf-cvpr12} formulated saliency estimation using two Gaussian filters by which color and position are respectively exploited to measure region uniqueness and variance of the spatial distribution. Yan {\em et al.}~\cite{hs-cvpr13} proposed a hierarchical framework that infers important values from three image layers in different scales. Also using a hierarchical indexing mechanism, Cheng {\em et al.}~\cite{gc-iccv13} proposed a Gaussian Mixture Model based abstract representation which decomposes an image into large scale perceptually homogeneous elements. But their saliency cues integration based on the compactness measure may not always be effective. Typical limitations of the existing methods based on contrast priors include attenuated object interior and ambiguous saliency detection for images with rich structures in foreground or/and background.

Complementing the prime role of contrast priors in this research topic, background priors~\cite{gs-eccv12} have been proposed recently to exploit two interesting priors about backgrounds -- connectivity and boundary priors. The background prior is based on an observation that the distance of a pair of background regions is shorter than that of a region from the salient object and a region from the background. Wei {\em et al.}~\cite{gs-eccv12} exploited background priors and the geodesic distance for the saliency detection. Yang {\em et al}.~\cite{GM-cvpr13} proposed a graph-based manifold ranking approach to characterize the overall differences between salient objects and background. Jiang {\em et al}.~\cite{MC-iccv13} integrated the background cues into the designed absorbing Markov chain. Regarding image boundaries as likely cues for background templates, Li {\em et al}.~\cite{DSR-iccv13} proposed a saliency detection algorithm from the perspective of dense and sparse appearance model reconstructions. However, these methods fail when objects touch the image boundary to quite some extent, or when connectivity assumptions are invalid in the presence of complex backgrounds or textured scenes. For instance, the maple leave case in Fig.~\ref{fig:fig1} poses a challenge for the method~\cite{gs-eccv12}.

Energy minimization based methods have also been introduced for saliency detection. Liu {\em et al}.~\cite{KDE-icip2010} proposed a nonparametric saliency model based on kernel density estimation (KDE). Jiang {\em et al}.~\cite{CB-bmvc11} proposed an iterative energy minimization framework to integrate both bottom-up salient stimuli and an object-level shape prior. Treating saliency computation as a regression problem, Jiang {\em et al}.~\cite{DRFI-cvpr13} integrated regional contrast, regional property and regional backgroundness. Chang {\em et al}.~\cite{GOVS-iccv11} proposed to account for the relationships of objectness and saliency by iteratively optimizing an energy function. 

This paper provides a more complete understanding of the PISA algorithm first presented in the conference version~\cite{pisa-cvpr13}, giving further background, insights, analysis, and evaluation. Furthermore, we improve the previous framework in two aspects. First, the improved PISA is cast as the energy minimization problem, which efficiently solved by the edge-aware cost-volume filter to generate the spatially coherent and fine-grained saliency maps in one shot. Second, for suppressing the effect of background, a more general spatial prior is integrated in our framework to obtain more compact saliency maps.

\section{Problem Formulation}
\label{sec:formul}

In this section, we introduce the formulation of PISA, and briefly overview the main components. 

Given an input image $I$, the objective of PISA is to extract salient objects automatically and assign consistently high saliency levels to them. Without loss of generality, we achieve this goal by minimizing the following energy function

\vspace{-6pt}
\begin{equation}
\label{equ:energyEqu}
E = \sum_{ p \in I } A(S_p) + C(S_p),			
\vspace{-2pt}    	
\end{equation}
where $A(S_p)$ represents the cost of labeling pixel $p$ with the saliency level $S_p$, which composes the data term according to the contrast based measures. $C(S_p)$ defines the neighborhood coherence to preserve the local structures and edges centered at $p$. We further specify $A(S_p)$ as

\begin{figure}[hptb]
\begin{center}
\includegraphics[width= \columnwidth]{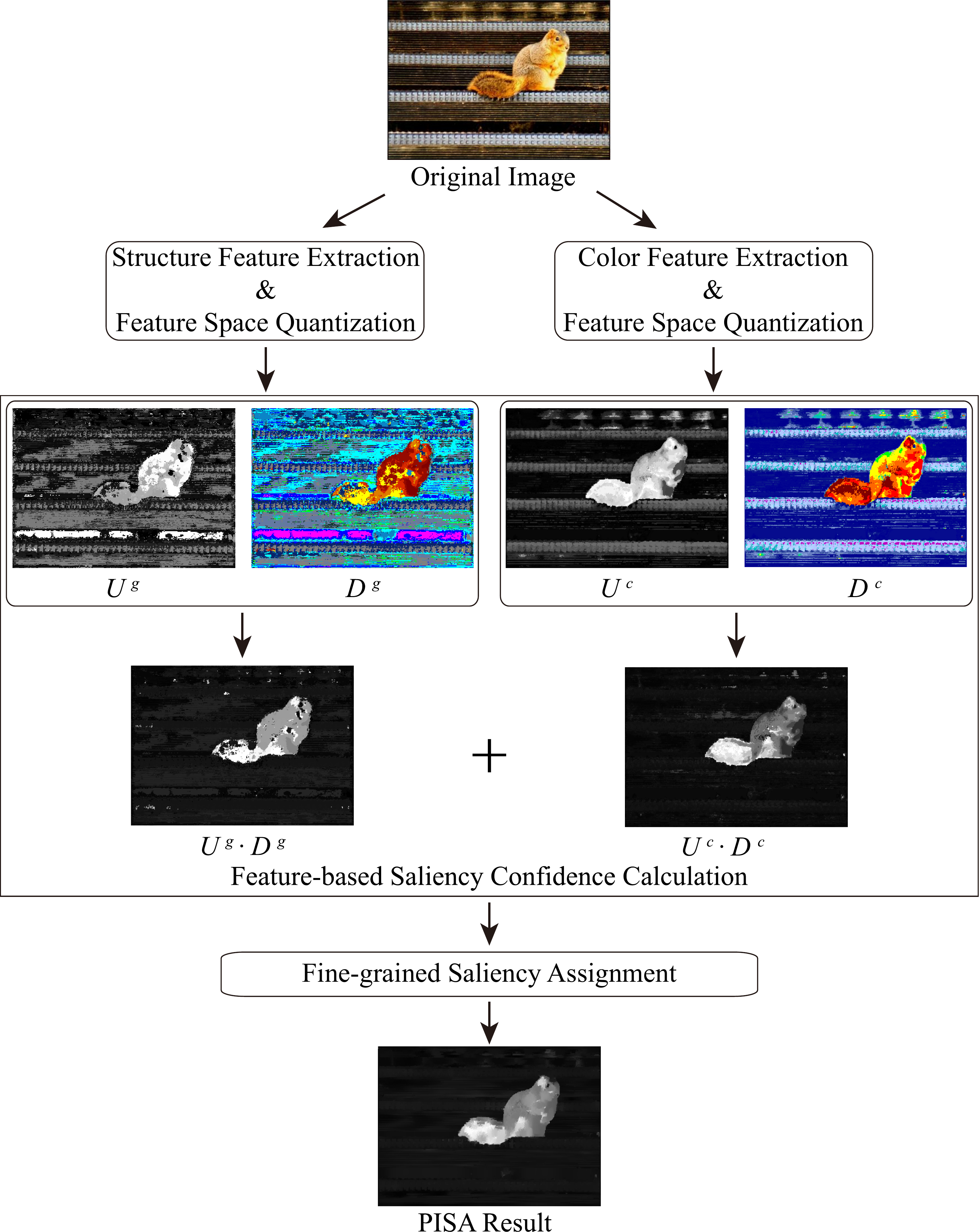}
\end{center}
\vspace{-11pt}
\caption{The main flowchart of PISA. The original image is on the top, $U^c$/$U^g$ denotes the color/structure contrast measure, and $D^c$/$D^g$ denotes the spatial prior term corresponding to the certain feature measure.}
\vspace{-12pt}
\label{fig:framework}
\end{figure}

\vspace{-5pt}
\begin{equation}
\label{equ:datacost}
A(S_p) = \Arrowvert S_p - f(p) \Arrowvert _2^2,
\end{equation}
where $f(p)$ denotes the normalized feature measure of $p$, aggregating two complementary contrast measures defined in a global context.  Fig.~\ref{fig:framework} illustrates the main flowchart of PISA.

\subsection{Feature-Based Saliency Confidence}

We introduce two types of features to capture contrast information of salient objects with respect to the scene background. They are a color-based contrast feature and a structure-based contrast feature, each of which is further integrated with the spatial priors holistically. These two features complement each other in detecting saliency cues from different perspectives, and are combined together in a pixelwise adaptive manner to measure the saliency. More formally, given an image $I$, we compute the feature-based saliency confidence $\hat{f}(p)$ for each pixel $p$ by aggregating the two contrast measures (i.e. the uniqueness in the feature spaces) $\{U^c(p), U^g(p)\}$ with the spatial priors $\{D^c(p), D^g(p)\}$, as

\vspace{-8pt}
\begin{equation}
\label{equ:contrast}
\hat{f}(p) = U^c(p) \cdot D^c(p) + U^g(p) \cdot D^g(p).
\vspace{-2pt}
\end{equation}

{\bf Appearance contrast term} $\{U^c(p), U^g(p)\}$.
The contrast measure is proposed based on the observation or principle that rare or infrequent visual features in a global context give rise to high salient values~\cite{hcrc-cvpr11, sf-cvpr12, gs-eccv12}. Here we exploit the structure-based contrast measure in addition to the well exploited color-based contrast measure, and we fuse the two measures $\{U^c(p), U^g(p)\}$ to achieve better performance. $U^c(p)$ denotes the uniqueness of pixel $p$ with respect to the entire image in the color feature space, and $U^g(p)$ denotes the uniqueness of pixel $p$ in the orientation-magnitude (OM) feature space. Their detailed implementations will be discussed in Sect.~\ref{sec:cc} and Sect.~\ref{sec:sc}, respectively. Instead of describing the features for pixel $p$ via its assigned superpixel, we use the non-parametric histogram distribution to capture and represent both the color and structure features with an appropriate observation region around $p$. It is worth mentioning that our framework is very general to incorporate more saliency cues in the similar way.

{\bf Spatial priors term $\{D^c(p), D^g(p)\}$}.
They are evaluated based on the generally valid spatial prior that salient pixels tend to distribute near the image center and away from the image boundary, i.e. people tend to frame an image by placing salient objects of interest in the center with background borders. Thus, we integrate the image center preference and boundary exclusion in the saliency reweighting process. We use $D^c(p)$ and $D^g(p)$ to denote the integration of image center spatial distance and image boundary exclusion of visually similar peers on the color and structure contrast measurement, respectively (Sect.~\ref{sec:spatial}). After reweighting the above saliency measurement based on appearance contrast, we keep the salient pixels compact and centered with the exclusion to the image boundary in the image domain.

We normalize the feature-based saliency confidence to the discrete saliency level set $\{0, 1, ...,\mathcal{L}-1\}$ for further calculating the label cost $A(S_p)$. This normalization is given by the following sigmoid-like function:

\vspace{-8pt}
\begin{equation}
\label{equ:sigmoid}
f(p) = R( \frac{\mathcal{L}-1}{1 + \exp(-\hat{f}(p))} ),
\vspace{-3pt}
\end{equation}
where $R$ denotes a rounding function, which rounds a float-point number to the nearest integer, and $\mathcal{L} - 1$ is the user defined maximum saliency level. We fix $\mathcal{L}$ to 24 in our all experiments.

\subsection{Coherence Constraint}
\label{sec:coherence}

To suppress spurious noises and non-uniform saliency assignment, we further incorporate the spatial connectivity and correlation constraint among neighborhood pixels together with the feature-based measures. The saliency level $S_p$ for pixel $p$ should be consistent with its neighborhood pixels which have similar appearance with $p$ within its local observation region $\Omega_p$ in the image domain. The coherence constraint $C(S_p)$ can be thus defined as

\vspace{-5pt}
\begin{equation}
\label{straint}
C(S_p) = \sum_{q \in \Omega_p}{\omega_{pq} \Arrowvert S_p-S_q \Arrowvert _2^2},
\end{equation}
where the observation window $\Omega_p$ for the anchor pixel $p$ delineates an arbitrarily-shaped and connected local support region (see Fig.~\ref{fig:feature}), $q$ represents a neighboring pixel to $p$ in $\Omega_p$, and $S_q$ is the saliency level assigned to $q$. $\omega_{pq}$ encodes the similarities between $p$ and $q$ within $\Omega_p$, which will be explained in the next section.


\section{Proposed Approach}
\label{sec:alg}

In this section, we unfold the framework of PISA and discuss the implementation details. In addition, a faster version of PISA, namely F-PISA, is also developed to greatly improve the runtime efficiency and keep comparable performance.

\begin{figure} [!t]
   \begin{center}
   \includegraphics[width = 0.75 \columnwidth]{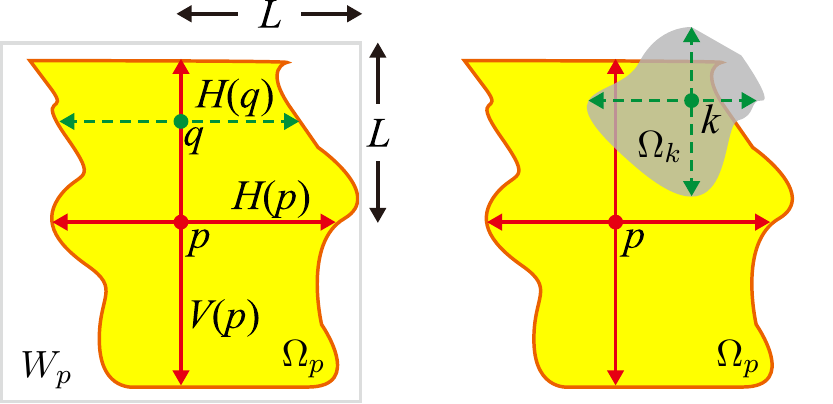}
   \end{center}
   \vspace{-10pt}
   \caption
   {\label{fig:clmf} Illustration of generating a pixelwise adaptive observation region~\cite{clmf-cvpr12}. The left subfigure shows the construction of the shape-adaptive observation window for pixel $p$, and the right subfigure shows the weighted aggregation of each $k\in\Omega_p$. }
   \vspace{-15pt}
   \end{figure}

\subsection{Pixelwise Adaptive Observation}
\label{sec:observationwindow}

Unlike the traditional methods\cite{hcrc-cvpr11,salobj-iccv11} that usually process fixed-size windows or over-segmented superpixels, PISA computes saliency by generating an arbitrarily-shaped observation region for each pixel in the image. This pixelwise observation plays a key role in feature extraction and fine-grained saliency assignment.

For a pixel $p$ centered at a square window $W_p$, we first define a color similarity criterion for a test pixel $q$ as follows,

\vspace{-5pt}
\begin{equation}
\label{equ:adaptive}
\begin{gathered}
|I_c(q) - I_c(p)| \le \tau,~c \in \{R,G,B\},~ q \in W_p,
\end{gathered}
\end{equation}
where $I_c$ is the intensity of the color band $c$ of the $3 \times 3$ median smoothed input image $I$. Set empirically, $L$ denotes the preset maximum arm length of the observation window $W_p$ centered at pixel $p$ (the size of $W_p$ is $(2L+1)\times(2L+1)$), and $\tau$ controls the confidence level of the color similarity. The method of generating $\Omega_p$ follows our previous study in image filtering (i.e. Cross-based Local Multipoint Filtering)~\cite{clmf-cvpr12}. We first decide a pixelwise adaptive cross with four arms (left, right, up, bottom) for every pixel $p$. By changing four arms of every pixel $p$ adaptively, the local image structure is captured reliably. These arms record the largest left/right horizontal and up/bottom vertical span of the anchor pixel $p$, where all the pixels covered by the arms are similar to pixel $p$ in color (i.e. they satisfy Eqn.~(\ref{equ:adaptive})). Let $H(p)$ and $V(p)$ denote all the pixels covered by the horizontal and vertical arms of the pixel $p$, respectively. Let $q$ denote any pixel covered by the vertical arms of the pixel $p$ (i.e. $q \in V(p)$), as shown in Fig. \ref{fig:clmf}. Then we can further construct the arbitrarily-shaped, connected local observation window $\Omega_p$ by integrating multiple $H(q)$ sliding along $V(p)$

\vspace{-12pt}
\begin{equation}
\Omega_p = \bigcup_{q \in V(p)}{H(q)}.
\end{equation}

\begin{figure} [!t]
   \begin{center}
   \includegraphics[width = \columnwidth]{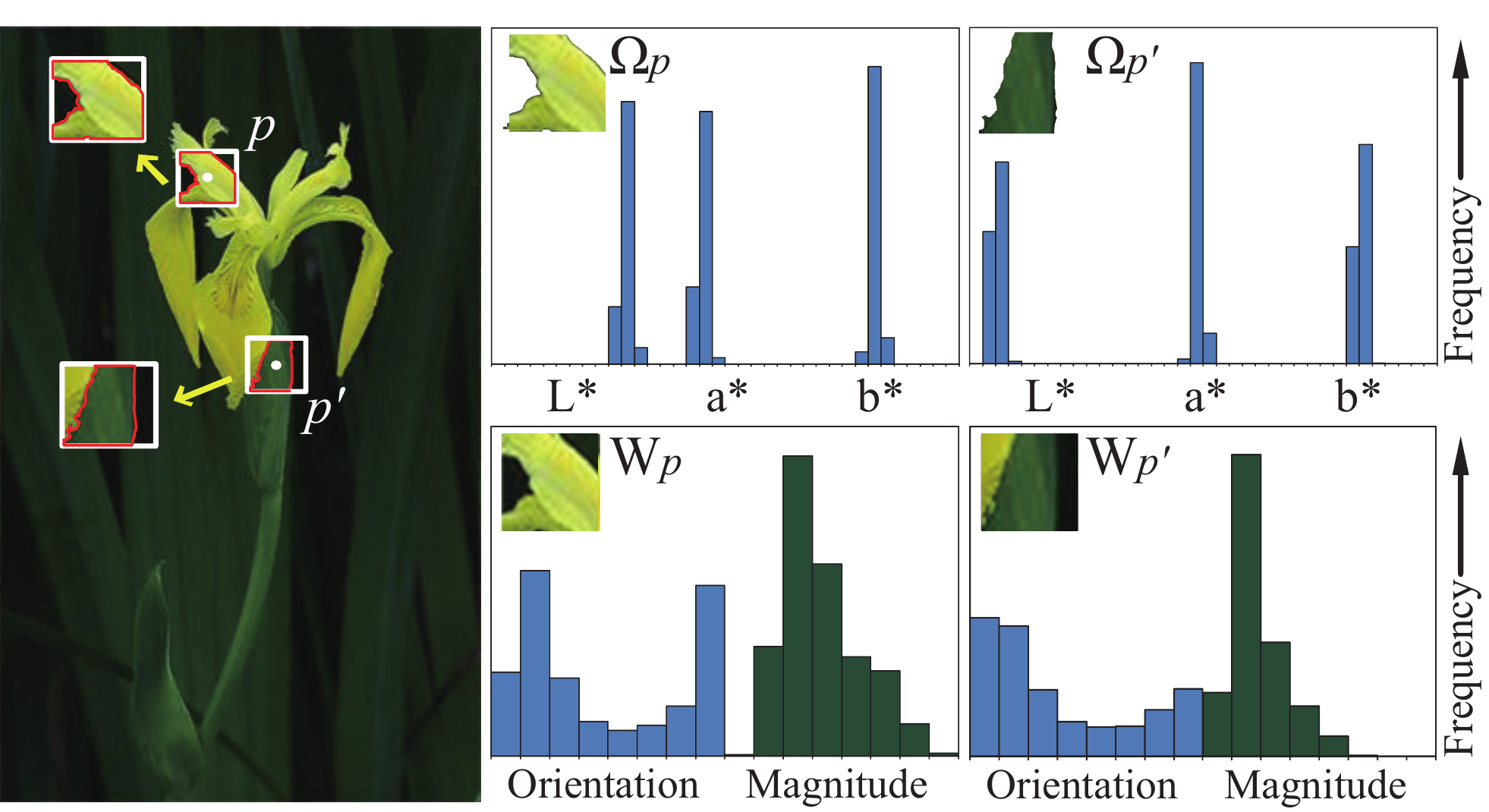}
   \end{center}
   \vspace{-10pt}
   \caption{\label{fig:feature} The color descriptor is extracted from the shape-adaptive region $\Omega_{p/p'}$ (top) and the orientation-magnitude (OM) descriptor captures the structures within a local window $W_{p/p'}$ (bottom). }
\vspace{-15pt}
\end{figure}

\subsection{Color and Structure-Based Saliency Measures}
\label{sec:fp}
\subsubsection{Color-Based Contrast}
\label{sec:cc}

Directly computing pixelwise color contrast in a global image context is computationally expensive, as its complexity is $O$($N^2$) with $N$ being the number of pixels in the image $I$. Recently, Cheng {\em et al.}\cite{hcrc-cvpr11} proposed an effective and efficient color-based contrast measure, i.e., histogram-based contrast (HC). They assume that if neglecting spatial correlations, pixels with the similar color value should have the same saliency value. However, without taking the neighborhood of pixels into consideration, their strategy of defining contrast on color information of individual pixels is sensitive to noise, and it is not extensible for measuring additional attributes. In this work, we compute the color contrast based on a non-parametric color distribution extracted from a local homogeneous region. As pixels within the homogeneous region share similar appearance with the central pixel, it is more robust to define a contrast measure on color information of homogeneous regions rather than individual pixels.

For each pixel $p$, we first construct a local observation region efficiently as described in Sect.~\ref{sec:observationwindow}. A color histogram ${\bf h}^c(p)$ for pixel $p$ is then built from the pixels $q\in\Omega_p$ covered in the localized homogeneous region. Using ${\bf h}^c(p)$ rather than $I_p$ is more consistent with psychological evidences on human eyes' receptive field on homogeneous regions. Using the $Lab$ color space, we quantize each color channel uniformly into 12 bins, so the color histogram ${\bf h}^c(p)$ is a 36-d descriptor (see Fig.~\ref{fig:feature}).

Next, we cluster pixels that share similar color histograms together using {\em kmeans}. The whole color feature space for the input image $I$ is then quantized into $K_c$ clusters, indexed by $\{ \phi_1, \dots, \phi_{K_c} \}$. As a result, we use the rarity of color clusters as the proxy to evaluate the rarity or contrast measure of pixels. Let $\phi_p$ denote the cluster that pixel $p$, or more precisely ${\bf h}^c(p)$, is assigned to. We estimate the color-based contrast measure $U^c(p)$ for pixel $p$ as

\begin{figure} [!t]
\begin{center}
\includegraphics[width = \columnwidth]{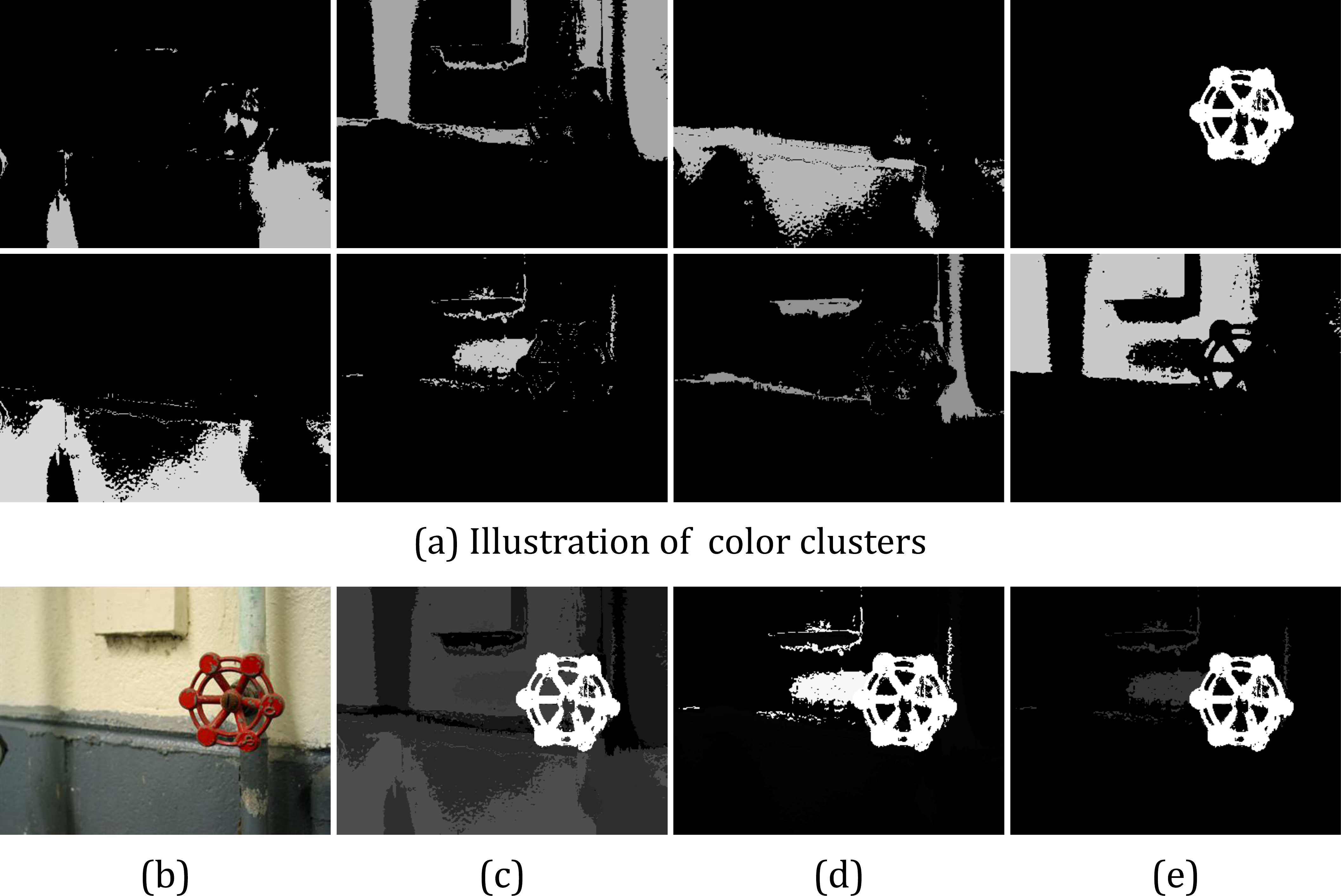}
\end{center}
\vspace{-10pt}
\caption
{\label{fig:spatial} Color-based contrast measure. (a) Assigning the pixels into eight clusters in the color feature space. (b) The input image from~\cite{ft-cvpr09}. (c) Color contrast measure $U^c$. (d) Spatial prior $D^c$. (e) Spatial prior-modulated color measure $U^c \cdot D^c$.  }
\vspace{-15pt}
\end{figure}

\vspace{-18pt}

\begin{equation}
	\label{equ:rarityc}
    U^c(p) = U^c({\bf h}^c(p)) = \sum_{i=1}^{K_c} \omega_i \Arrowvert {\bf h}^c(\phi_i), {\bf h}^c(\phi_p) \Arrowvert,
\vspace{-2pt}
\end{equation}
where $\omega_i$ uses the number of pixels belonging to the cluster $\phi_i$ as a weight to emphasize the color contrast to bigger clusters, and ${\bf h}^c(\phi_p)$ is the average color histogram of cluster $\phi_p$. Fig.~\ref{fig:spatial} (a) illustrates an example image with eight color clusters and their contrast measure $U^c(p)$.

Feature space quantization may cause undesirable artifacts. When directly calculating the $L_2$ distance of histograms or giving an inappropriate cluster number $K_c$, similar color histograms can sometimes be quantized into different clusters. We tackle this problem in three aspects: i) Improve clustering with color dissimilarity. We sightly modify {\it kmeans} in its distance when clustering. In addition to the $L_2$ distance between the two histograms, we add the color dissimilarity between the center pixels into the distance measurement. ii) Decide $K_c$ adaptively according to the histogram distribution. The cluster number $K_c$ of the color feature space is adaptively decided with regard to the image content. Similar to that used in~\cite{hcrc-cvpr11}, we choose the most frequently occurring color features by ensuring they cover 95\% of the histogram distributions of all pixels in the input image $I$. iii) Reweight the salient values of clusters with respect to their visual similarities. We adopt a linearly-varying smoothing scheme~\cite{hcrc-cvpr11} to refine the quantization-based saliency measurement. The saliency value of each cluster is replaced by the weighted average of the saliency values of visually similar clusters. Larger weights are assigned to those clusters which share similar color features. Such a refinement smooths the saliency assignments to each pixel.

Our proposed method that computes the color contrast based on non-parametric color distribution reduces the computational complexity from $O(N^2)$ to $O(N\cdot K_c) + O(K_c^2)$, where the second term corresponds to the complexity of {\em kmeans} and usually is very small. As we observed, $K_c$ typically takes values in the range of 6 to 403 in the ASD dataset \cite{ft-cvpr09} which contains 1000 images.

\subsubsection{Structure-Based Contrast}
\label{sec:sc}
As discussed in Sect. \ref{sec:intro}, using only color information is not adequate to completely depict salient objects or parts of them against the non-salient background. Even though in the cases that the color-based measure produces good results, other complementary measures can still contribute to reinforce the saliency assignment. Therefore, we propose a structure-based measure to complement the color-based contrast measure here.
The proposed structure-based measure models the image gradient distribution for every pixel $p$ by a histogram ${\bf h}^g(p)$ in a rectangular region $W_p$. ${\bf h}^g(p)$ measures the occurrence frequency of a concatenated vector consisting of a gradient orientation component and a gradient magnitude component. Similarly, we quantize both components into eight bins, and call the resulting feature space the OM space. It is clear that a point in such a OM space is 16-d (see Fig.~\ref{fig:feature}). In this paper, we fix the local window $W_p$ to the same size as the maximum observation window of the color histogram extraction for the comparability. As will be shown later, we find that our OM structure descriptor, though simple, is more effective and reliable than other gradient features such as Gabor\cite{gabor-pami96} and LBP\cite{lbp-pr09} in the image saliency detection task.

Similar to the color contrast measure, {\em kmeans} is utilized to partition the OM feature space into $K_g$ clusters, indexed by $\{ \varphi_1, \dots, \varphi_{K_g} \}$. The structure contrast measure for pixel $p$ is equivalent to measuring that of the cluster $\varphi_p$ which $p$ is grouped to as

\vspace{-18pt}
\begin{equation}
\label{equ:rarityg}
U^g(p) = U^g({\bf h}^g(p)) = \sum_{i=1}^{K_g} \omega_i \Arrowvert {\bf h}^g(\varphi_i), {\bf h}^g(\varphi_p) \Arrowvert,
\end{equation}
where $\omega_i$ is the weight stressing the contrast against bigger clusters, and ${\bf h}^g(\varphi_p)$ is the average OM histogram of the cluster $\varphi_p$.

$U^g$ may suffer from the influence of side effects caused by the brute-force feature space quantization process.  Again, we alleviate these artifacts by adopting the same strategy illustrated in Sect.~\ref{sec:cc}. i.e., using slightly modified {\it kmeans}, determining the cluster number $K_g$ adaptively by representing the most frequent OM vectors and accounting for at least 95\% pixels, and applying local smoothing scheme. We observe $K_g$ typically varies from 11 to 43 in the ASD dataset~\cite{ft-cvpr09}.

\subsubsection{Spatial Priors}
\label{sec:spatial}
Motivated by recent works\cite{sf-cvpr12, ca-cvpr10, l2d-pami11, GM-cvpr13}, we impose a spatial prior term on each of the two contrast measures $\{U^c(p), U^g(p)\}$, constraining pixels rendered salient to be centered and excluded to the image boundary in the image domain based on the image center preference and the image boundary exclusion. 
For each pixel $p$, we evaluate the initial spatial prior term $\tilde{D}^{c/g}{(p)}$ based on the cluster $\phi_i/\varphi_i$ that contains $p$ from two aspects: i) preference to the image center, and ii) exclusion to the image boundary. Combining these two criteria, we compute $\tilde{D}^{c/g}{(p)}$ as follows:

\vspace{-8pt}
\begin{equation}
\label{equ:spatial}
\tilde{D}^{c/g}(p) = \sum_{l=1}^{n_p} ( \frac{  \Arrowvert {\bf x}_l, {\bf {c}} \Arrowvert^2 } {n_p} + \lambda \cdot \frac {{\bf 1}_{\Lambda}({\bf x}_l)}{|\Lambda|} ) \;,
\end{equation}
where $n_p$ is the number of pixels which are contained in the same color (or OM) cluster $\phi_i$ (or $\varphi_i$) with $p$. ${\bf {c}}$ is the image center position. $\Lambda$ indicates the image border region, which is formed by the pixels close to the image borders. As a matter of fact, this region typically belongs to non-salient background. Thus, we incorporate the $\frac{ \sum{{\bf 1}_{\Lambda}(\cdot)}} {| \Lambda |}$ as the probability of $\phi_i$/$\varphi_i$ belonging to the image border region. We use a user-specified parameter $\lambda$ to control the relative weight of the image boundary exclusion. Fig. ~\ref{fig:spatial} (d) and (e) illustrate the spatial prior together with using the color-based measure and the effectiveness for saliency assignment.

Since clusters closer to the image border or farther from the image center are often unlikely to be salient, we compute the final spatial prior term $D^{c/g}(p)$ for pixel $p$ using a threshold $\mathcal{T}$ as

\vspace{-8pt}
\begin{equation}
\label{equ:decision}
D^{c/g}(p) =
\begin{cases}
\exp(- \kappa \cdot \tilde{D}^{c/g}(p)); & \tilde{D}^{c/g}(p) \leq\mathcal{T} \\
0; & \text{otherwise} \;.
\end{cases}
\end{equation}
where $\kappa$ controls the fall-off rate of the exponential function.
By now we have defined all the four terms necessary for computing $\hat{f}(p)$ in Eqn. (\ref{equ:contrast}).

\subsection{Fine-Grained Saliency Assignment}
\label{sec:optimize}

Our goal is to assign each pixel $p$ in the image $I$ to a saliency level $\mathcal{S}$ from the discrete saliency level set $\{0, 1, ..., \mathcal{L} - 1\}$, with the formulation in Eqn.~(\ref{equ:energyEqu}). This is a multi-labeling minimization task integrating a data term and a smoothness term. Instead of using global discrete optimization methods, we employ the cost-volume filtering technique~\cite{fcv-pami13} to achieve this goal, which computes the discrete assignment efficiently while keeping local labeling coherence. Specifically, this method
aggregates the label costs within a support window by applying a local edge-preserving smoothing filter, and then selects the label in a Winner-Takes-All fashion. The fine-grained saliency is computed for each pixel with the following steps.

\textbf{(i) Constructing the cost-volume}: Following~\cite{fcv-pami13}, the cost-volume is a three dimensional array, and each element $V_{p, \mathcal{S}}$ in the array represents the cost for choosing a saliency level $\mathcal{S}$ at pixel $p$. We compute $V_{p, \mathcal{S}}$ as the square difference between $\mathcal{S}$ and the normalized feature-based saliency measure $f(p)$:

\vspace{-8pt}
\begin{equation}
\label{equ:costvolume}
V_{p, \mathcal{S}} = \Arrowvert \mathcal{S} - f(p) \Arrowvert _2^2.
\end{equation}

\textbf{(ii) Filtering the cost-volume}:
To smooth the label costs in the image domain, the cost-volume will be further filtered with an edge-preserving filter. The original cost volume filtering method uses the guided filter~\cite{GF-pami13}, which employs fixed-sized square observation windows, and it derives the output of the filtering simply as an average of multiple linear regression results from shifted windows of neighboring pixels.

In this work, to incorporate the local edge-aware coherence (Eqn.~(\ref{straint})) and also to achieve more efficient runtime, we extend the guided filter into a new form based on the pixelwise adaptive observation~\cite{clmf-cvpr12}. Specifically, for pixel $p$ we estimate the correlation of $p$ and its neighbor $k \in \Omega_p$ by 

\begin{equation}
\omega_{p,k}=\frac{|\Omega_k|}{\sum_{k \in \Omega_p}|\Omega_k|},
\end{equation}
where $\Omega_k$ is the observation region of pixel $k$ (a neighbor of $p$), and $|\Omega_k|$ denotes the number of pixels in $\Omega_k$. Intuitively, the correlation of pixel $p$ and the neighbor $k$ is proportional to $|\Omega_k|$.  We refer to~\cite{clmf-cvpr12} for the technical background. The cost of $p$ can be updated by the weighted average of the initial costs of all pixels in $\Omega_p$ as

\begin{equation}
\label{equ:filter}
V_{p, \mathcal{S}}^{'} = \sum_{k \in \Omega_p} \omega_{p,k} V_{k, \mathcal{S}}. 
\end{equation}

This step encourages the saliency values to be smooth in the homogeneous regions and also preserves the object details (e.g. edges and structures) in  the fine-grained saliency assignment.

\textbf{(iii) Winner-Takes-All label selection}: After the cost-volume is updated, the final saliency level $S_p$ at pixel $p$ is selected by

\vspace{-12pt}
\begin{equation}
\label{equ:wta}
S_p = \arg \min_{\mathcal{S}} V_{p, \mathcal{S}}'.
\end{equation}

\subsection{F-PISA: Fast Implementation}
\label{sec:fast}

Salient object detection is always cast as a preprocessing technique for subsequent applications, which demands a fast and accurate solution. To optimize accuracy-complexity trade-off, we present a faster version F-PISA, which contains well-designed algorithmic choices. Instead of processing the full image grid, we perform a gradient-driven subsampling of the input image $I$, so the saliency computation in Eqn.~(\ref{equ:energyEqu}) is only applied to this set of selected pixels. More specifically, for a given image $I$, we pick the pixel with the largest gradient magnitude from a 3$\times$3 rectangular patch on the regular image grid to form a sparse image $I^l$. The two proposed contrast saliency measures with edge-preserving coherence are then computed for $I^l$, giving a sparse saliency map $S^l$. To obtain a full-resolution saliency map $S$, we propagate the saliency values among pixels in the same pixel-adaptive observation region, as they share the similar appearance. This propagation scheme resembles the principle of joint bilateral upsampling~\cite{jbu-tog07}, using a high-resolution color image $I$ as a guidance to upsample a sparsely-valued solution map $S^l$. It can produce a smoothly varying dense saliency map $S$ without blurring the edges of salient objects. Thus given a pixel $p \in I$, its saliency value is obtained as

\vspace{-5pt}
\begin{equation}
	\label{equ:fast}
		S(p) = \frac{1}{m}\sum^{m}_{i=1}w_{p,{k_i}}S^l(k_i),
\end{equation}
where $k_i$ belongs to $I^l$ and its pixel-adaptive support region $\Omega_{k_i}$ contains $p$, $m$ is the total number of such pixels, and $w_{p,{k_i}} = \exp(- \frac{\Arrowvert {{\bf x}_p, {\bf x}_{k_i}} \Arrowvert}{\sigma})$. In Sect.~\ref{sec:exper}, we evaluate the performance of this fast version quantitatively and qualitatively on six public benchmark datasets.

\section{Experiments}
\label{sec:exper}
We present empirical evaluation and analysis of the proposed PISA against several state-of-the-art methods (including the conference version~\cite{pisa-cvpr13}) on six public available datasets. We further analyze the effectiveness of the two complementary components, i.e., color-based contrast measure and structure-based contrast measure, as well as their corresponding spatial priors (image center preference and boundary exclusion). We justify the importance of the proposed energy minimization framework and the sigmoid-like function for the feature-based saliency confidence normalization. At last, we discuss our limitations through failure cases.

\subsection{Description of Datasets}

We evaluate the proposed methods on six public available datasets. They are ASD~\cite{ft-cvpr09}, SOD~\cite{sod-pocv10}, SED1~\cite{sed-cvpr07}, ECSSD~\cite{hs-cvpr13}, PASCAL-1500~\cite{LRR-bmvc13} and the Taobao Commodity Dataset (TCD)\footnote{http://vision.sysu.edu.cn/project/PISA/} newly created by us. The ASD is also called MSRA-1000 which contains 1000 images with accurate human-labeled masks for salient objects and has been widely used by recent methods. The SOD dataset is more challenging with complex objects and scenes included in its 300 images, and we obtain the ground-truth for this dataset from the authors of the work~\cite{gs-eccv12}. The SED1 dataset is exploited recently which contains 100 images of single objects, and we consider a pixel salient if it is annotated as salient by all subjects. The ECSSD contains 1000 diversified patterns in both background and foreground images, which includes many semantically meaningful but structurally complex images for evaluation. The PASCAL-1500, created from PASCAL VOC 2012, is also a challenge dataset, in which the images contain multiple objects appearing at a variety of locations and scales with cluttered background. The TCD dataset that we make available with this paper contains 800 commodity images from the shops on the Taobao website. The ground truth masks of the TCD dataset are obtained by inviting common sellers of Taobao website to annotate their commodities, i.e., masking salient objects that they want to show from their exhibition. These images include all kinds of commodity with and without human models, thus having complex backgrounds and scenes with highly complex foregrounds.

\vspace{-10pt}
\subsection{Experimental Setup}
\label{sec:setup}
We choose the total saliency level $\mathcal{L}=24$. For the step of generating pixelwise adaptive observation, we set \{$\tau$, $L$\} = \{60, 10\} to extract color features and build saliency coherence support regions. We set \{$\lambda, \kappa, \delta, \mathcal{T}$\} = \{$2.5 \times 10^4$, 0.006, 0.001, 30\}. 
While for F-PISA, we set \{$\tau$, $L$\} = \{50, 5\} and \{$\lambda, \kappa, \delta, \mathcal{T}$\} = \{$2\times10^3$, 0.035, 0.001, 30\}. These parameters are fixed in all experiments for the six datasets.

We use (P)recision-(R)ecall curves (PR curves), $F_{0.3}$ metric and MAE to evaluate all the algorithms. Given the binarized saliency map via the threshold value from 0 to 255, precision means the ratio of the correctly assigned salient pixel number in relation to all the detected salient pixel number, and recall means the ratio of the correct salient pixel number in relation to the ground truth number. Different from (P)recision-(R)ecall curves using a fixed threshold for every image, the $F_{0.3}$ metric exploits an adaptive threshold of each image to perform the evaluation. The adaptive threshold is defined as

\vspace{-8pt}
\begin{equation}
T = \frac{2}{W \times H}\sum_{x=1}^{W}\sum_{y = 1}^{H}{S(x, y)},
\vspace{-5pt}
\end{equation}
where $W$ and $H$ denote the width and height of an image, respectively. The F-measure is defined as follows with the precision and recall of the above adaptive threshold:

\vspace{-8pt}
\begin{equation}
	\label{equ:FM}
		F_{\beta^2} = \frac{(1 + \beta^2) \cdot Precision \cdot Recall}{\beta^2 \cdot Precision + Recall},
\vspace{-5pt}
\end{equation}
where we set the $\beta^2$ = 0.3 to emphasize the precision as suggested in \cite{ft-cvpr09}. 
As pointed out in \cite{sf-cvpr12}, PR curves and $F_{0.3}$ metric are aimed at quantitative comparison, while Mean Absolute Error (MAE) are better than them for taking visual comparison into consideration to estimate dissimilarity between a saliency map $S$ and the ground truth $G$, which is defined as

\vspace{-8pt}
\begin{equation}
MAE = \frac{1}{|I|}\sum_p{|S_p - G_p|},
\end{equation}
where $|I|$ is the number of image pixels.

\subsection{Experimental Results and Comparisons}
\begin{figure*} [htbp]
\begin{center}
\includegraphics[width = 1 \textwidth]{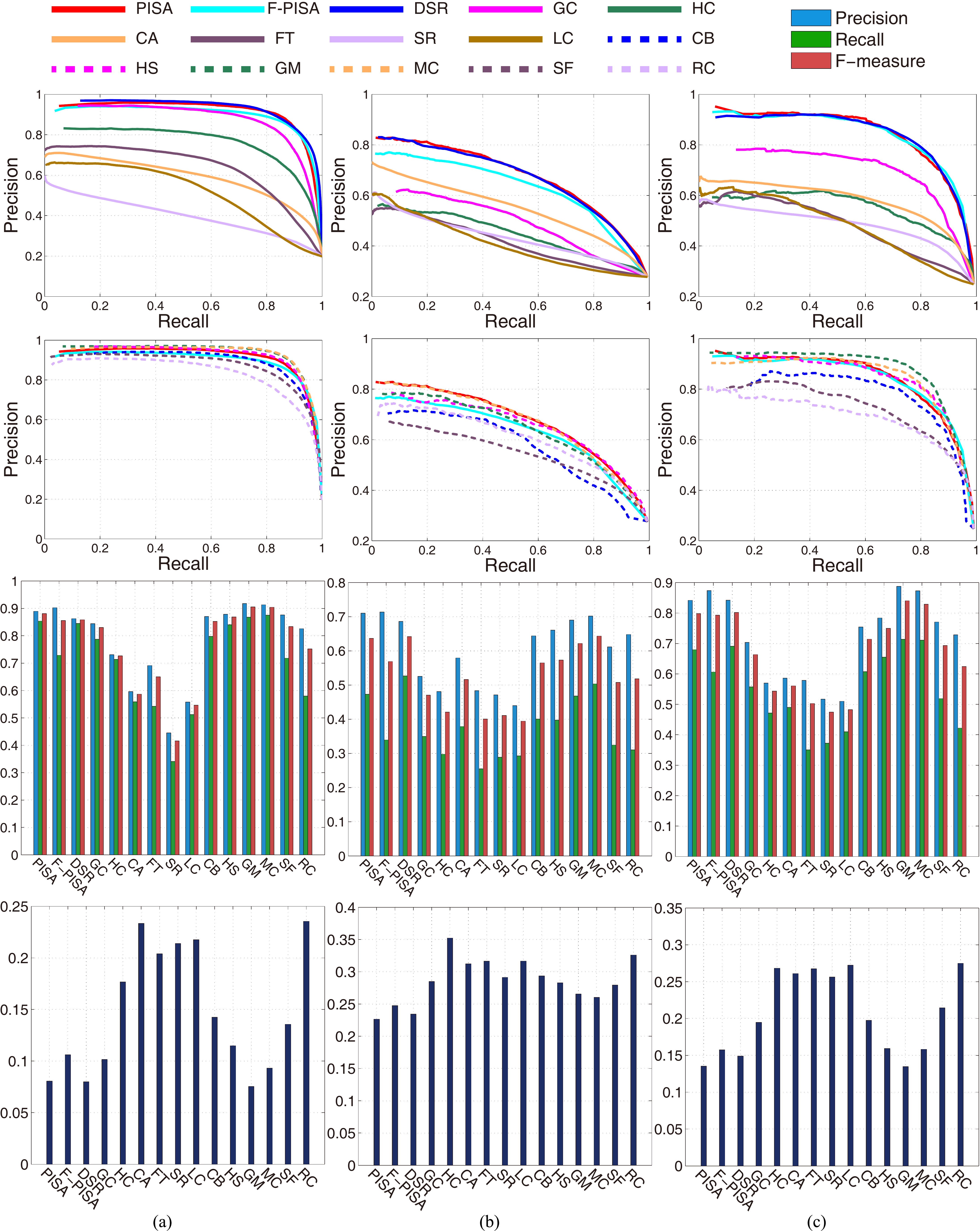}
\end{center}
\vspace{-10pt}
\caption
{\label{fig:PRF}
PR curves (first and second row), $F_{0.3}$ metric (third row) and MAE (fourth row) for comparing previous works with the proposed PISA and F-PISA methods on the three datasets from left to right: (a) ASD~\cite{ft-cvpr09}, (b) SOD~\cite{sod-pocv10}, (c) SED1~\cite{sed-cvpr07}, respectively. Our proposed methods PISA/F-PISA perform nearly the same as the state of art methods.}
\end{figure*}

\begin{figure*} [htbp]
\begin{center}
\includegraphics[width = 1 \textwidth]{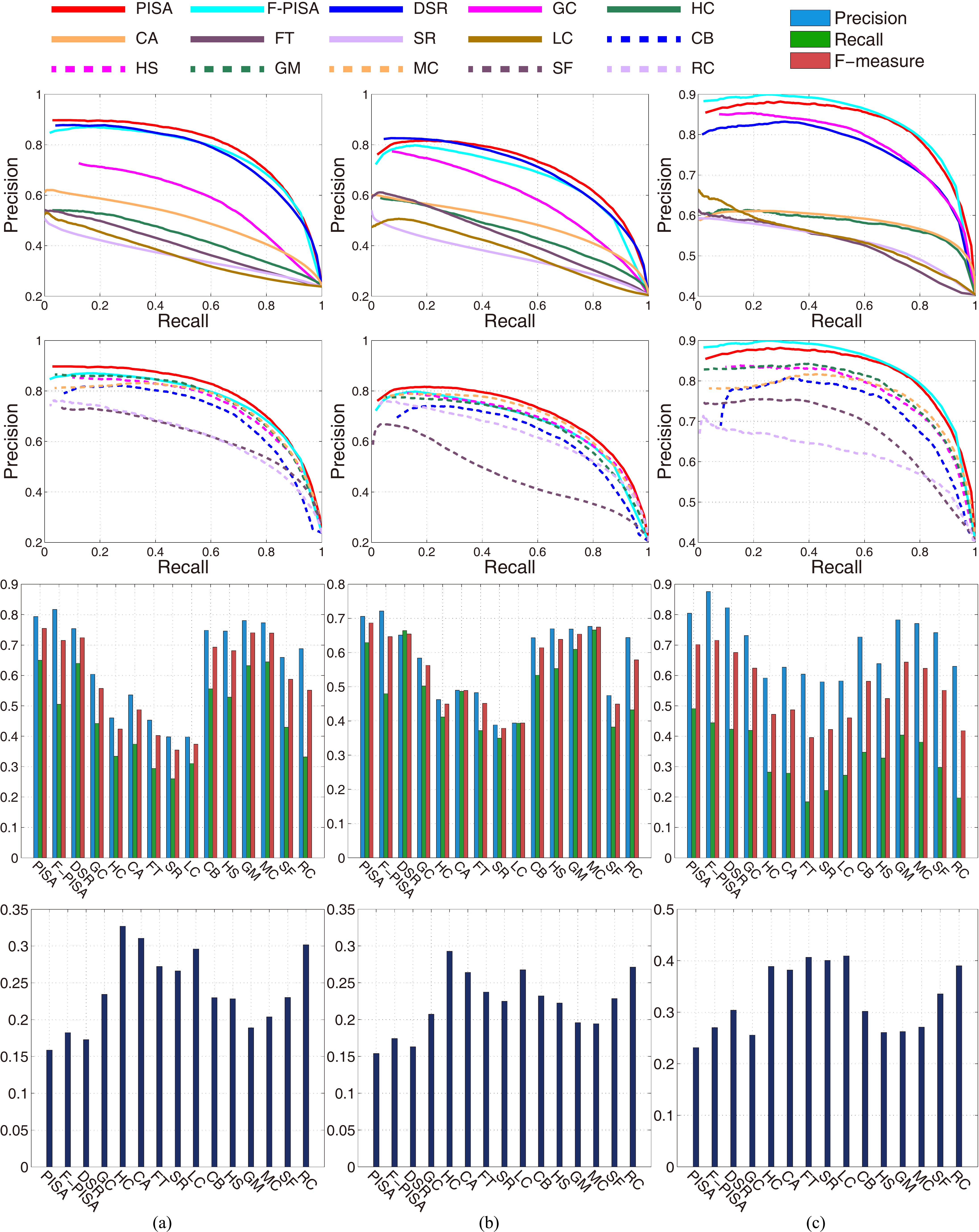}
\end{center}
\vspace{-10pt}
\caption
{\label{fig:PRF2}PR curves (first and second row), $F_{0.3}$ metric (third row) and MAE (fourth row) for comparing previous works with the proposed PISA and F-PISA methods on the three datasets from left to right: (a) ECSSD~\cite{hs-cvpr13}, (b) PASCAL-1500~\cite{LRR-bmvc13}, (c) proposed TCD, respectively. Our proposed methods PISA/F-PISA perform consistently better than the other methods.}
\end{figure*}

We compare our methods with thirteen recent state-of-the-art works: Dense and Sparse Reconstruction (DSR)~\cite{DSR-iccv13}, Global Cues (GC)~\cite{gc-iccv13}, Histogram-based Contrast (HC)~\cite{hcrc-cvpr11}, Context-Aware saliency (CA)~\cite{ca-cvpr10}, Frequency-Tuned saliency (FT)~\cite{ft-cvpr09}, Spectral Residual saliency~(SR)~\cite{sr-cvpr07}, Spatial-temporal Cues~(LC)~\cite{lc-cvpr06}, Context-Based saliency (CB)~\cite{CB-bmvc11}, Markov Chain saliency (MC)~\cite{MC-iccv13}, Hierarchical Saliency (HS)~\cite{hs-cvpr13}, Graph-based Manifold ranking (GM)~\cite{GM-cvpr13}, Saliency Filter (SF)~\cite{sf-cvpr12}, and Region-based Contrast (RC)~\cite{hcrc-cvpr11}. Whenever they are available, we use the author-provided results. Results of HC, FT, SR, LC, RC are generated by using the codes provided by~\cite{hcrc-cvpr11}, and we adopt the public implementations from the original authors for DSR, GC, CA, CB, HS, GM, MC and SF. Note that the saliency maps of all methods are mapped to the range [0, 255] by the same max-min normalization method for the further evaluation. The evaluation results are shown in Fig.~\ref{fig:PRF} and~\ref{fig:PRF2}, respectively. 


In Fig.~\ref{fig:PRF}, based on the PR curves of ASD, SOD and SED1, our proposed method PISA performs nearly the same as compared methods. To evaluate the overall performance of the PR curve, we calculate the average precision, which is the integral area under the PR curve. For the ASD dataset, our PISA, DSR, HS, GM and MC all achieve more than 93.0\% accuracy, while the average precision of PISA is 1.5\%, 0.6\%, 2.1\%, 1.7\% less than DSR, HS, GM, MC, respectively. For the SOD dataset, PISA, DSR and MC all achieve more than 80\% accuracy, while the average precision of PISA is 0.5\% better than both of them. For the SED1 dataset, PISA, DSR, HS, GM and MC all achieve more than 90.0\% accuracy, while the average precision of PISA is only 2.8\% less than GM. Based on the $F_{0.3}$ metric in Fig.~\ref{fig:PRF}, PISA obtains 2\% less than GM/MC on ASD, 0.5\% less than DSR/MC on SOD, 4\% less than GM/MC on SED1. Based on the MAE in Fig.~\ref{fig:PRF}, PISA obtains the best results on the SOD datasets and advances together with the best method GM on ASD/SED1. Hence, compared with all the compared methods, PISA is only slightly better on SOD, and is only a little worse on ASD and SED1. Since ASD and SED1 datasets are simple and not challenging, it is not suitable for showing the advantage of PISA.

The superior performance of PISA is demonstrated in Fig.~\ref{fig:PRF2}.
Based on the PR curves, $F_{0.3}$ and MAE in Fig.~\ref{fig:PRF2}, one can clearly see that our PISA consistently outperforms all the compared methods on ECSSD, PASCAL-1500, TCD, respectively. In particular, TCD is different in focusing on commodity images, whose salient objects contain diverse patterns and rich structure information. This is consistent with our motivations i) and ii) in Sect~\ref{sec:intro}. Designed to meet these objectives, our PISA achieves clearly higher performance than the compared methods. In addition, PISA in this paper performs 2\% better than the conference version (PISA-prev) on average, and readers are encouraged to see the supplementary file for more details.

\begin{figure*} [htbp]
\begin{center}
\hspace{-10pt}
\includegraphics[width = 0.85\textwidth]{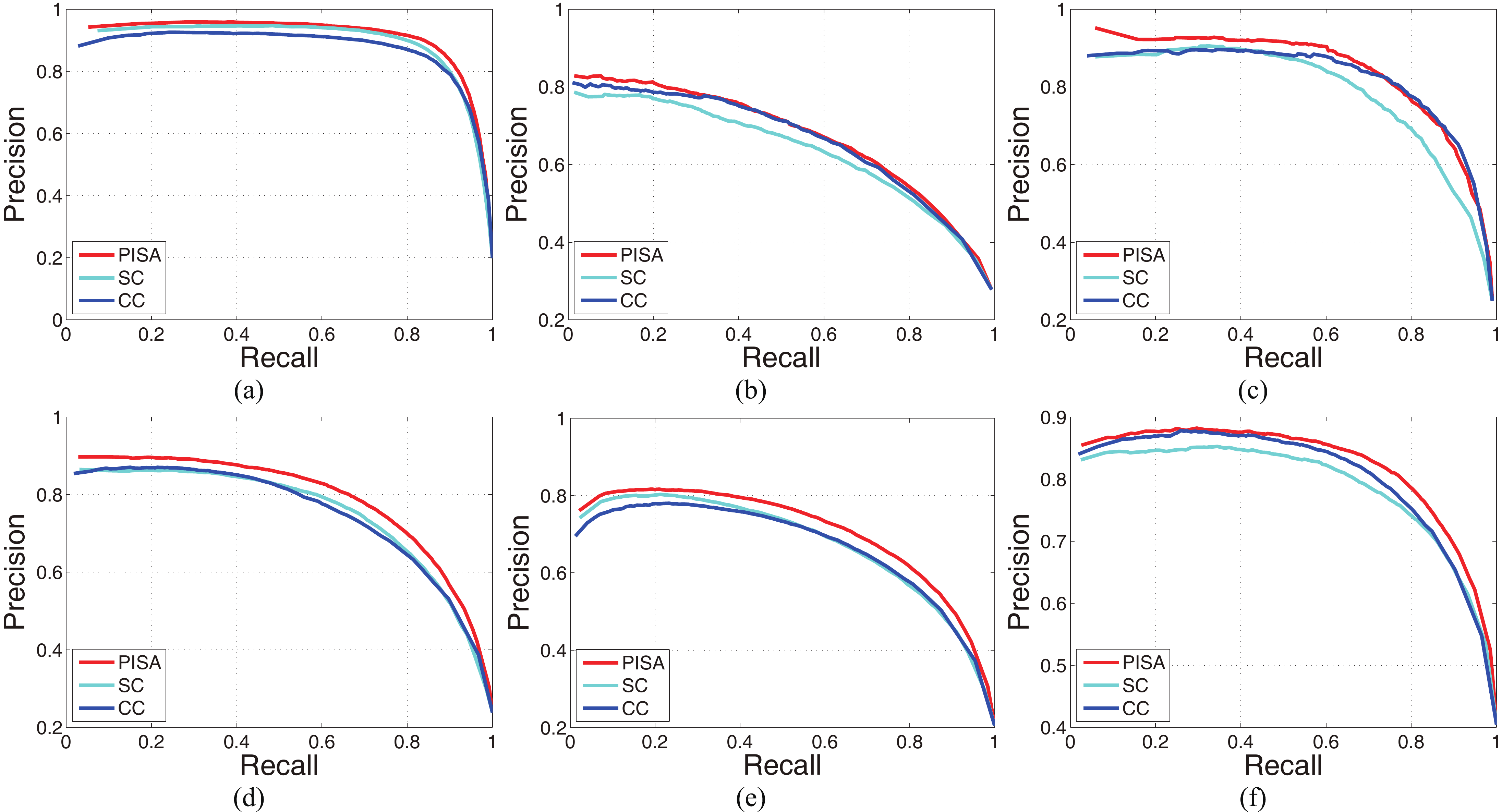}
\end{center}
\vspace{-15pt}
\caption
{\label{fig:cuePR} Extensive study for different saliency measures in our methods. The experiments are executed on all the six datasets, top row from left to right: ASD~\cite{ft-cvpr09}, SOD~\cite{sod-pocv10}, SED1~\cite{sed-cvpr07}, bottom row from left to right: ECSSD~\cite{hs-cvpr13}, PASCAL-1500~\cite{LRR-bmvc13}, TCD. We can observe the advantage of aggregating the two complementary contrast measures: structure-based contrast (SC) and color-based contrast (CC).}
\vspace{-10pt}
\end{figure*}

\begin{figure*} [htbp]
\begin{center}
\hspace{-10pt}
\includegraphics[width = 0.85 \textwidth]{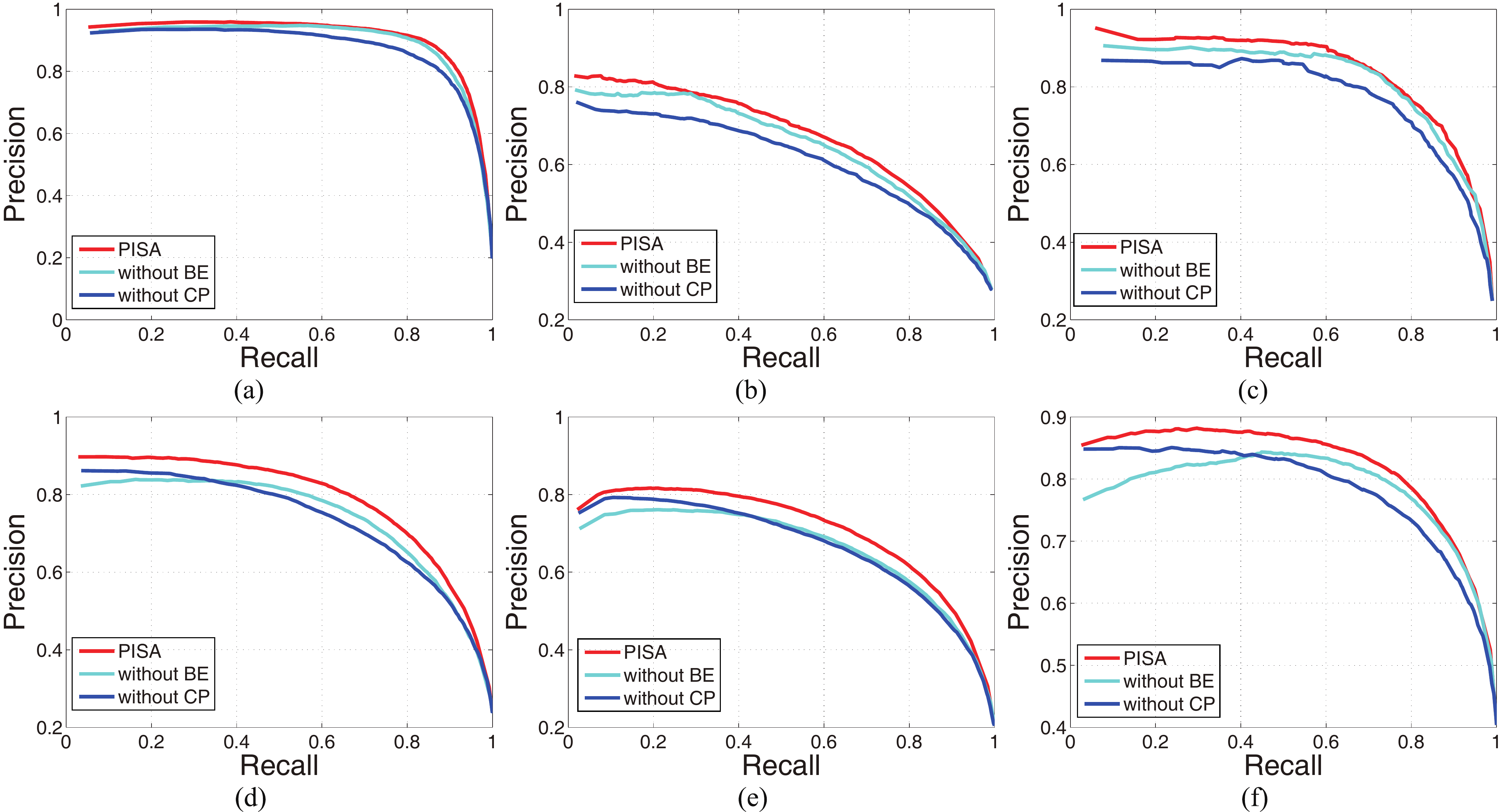}
\end{center}
\vspace{-15pt}
\caption
{\label{fig:compPR} Extensive study for the components of the proposed spatial prior in our methods. The experiments are executed on all the six datasets, top row from left to right: ASD~\cite{ft-cvpr09}, SOD~\cite{sod-pocv10}, SED1~\cite{sed-cvpr07}, bottom row from left to right: ECSSD~\cite{hs-cvpr13}, PASCAL-1500~\cite{LRR-bmvc13}, TCD. One can observe the contribution of image center preference (CP) and boundary exclusion (BE).}
\vspace{-10pt}
\end{figure*}

\subsection{Component Analysis}
\label{sec:component}
We further analyze the effectiveness of the two complementary measures, i.e. color-based contrast (CC) and structure-based contrast (SC). The quantitative results on the six datasets in Fig.~\ref{fig:cuePR} demonstrate the requisite of aggregating the two measures: PISA (SC + CC) performs consistently better than SC or CC alone.  
We can observe that the aggregated saliency detection achieves superior performance, as CC and SC capture saliency from different aspects, verified by the visual results in Fig.~\ref{fig:fig1}. It is worth noting that we obtain favorable results on the images in the second and third rows in Fig.~\ref{fig:fig1}, which are exhibited in~\cite{gs-eccv12} and~\cite{hcrc-cvpr11} as failure cases. They serve as good evidences to advocate our choice in fusing complementary saliency cues.

We also analyze the contribution of the introduced spatial priors, i.e. image center preference and boundary exclusion. The quantitative results on the six datasets in Fig.~\ref{fig:compPR} illustrate the advantage of introducing these spatial priors. ``without BE'' represents the PISA framework without boundary exclusion (BE) only, while ``without CP'' represents without image center preference (CP) only. Justified by the experiments on the six datasets, the introduced spatial priors contribute to achieve superior performance, as CP and BE represent the typical choices when people take pictures.

\begin{figure}[htbp]
\begin{center}
\includegraphics[width =1 \columnwidth]{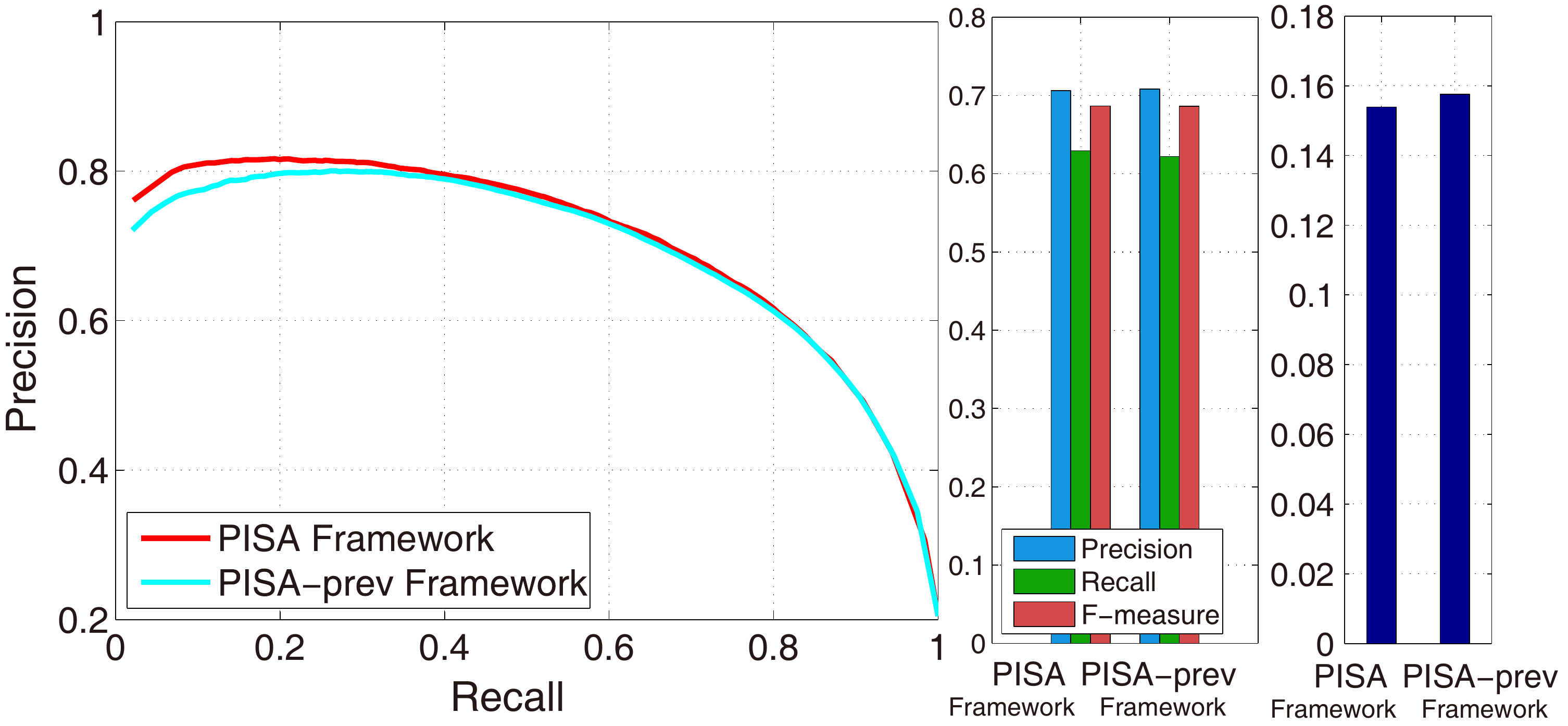}
\end{center}
\vspace{-15pt}
\caption
{\label{fig:compFramework} Empirical study on PASCAL-1500~\cite{LRR-bmvc13} for justifying the significance of the proposed framework, named ``PISA Framework'' . ``PISA-prev Framework'' denotes our conference version.}
\vspace{-10pt}
\end{figure}

\begin{figure}[htbp]
\begin{center}
\includegraphics[width = 0.91 \columnwidth]{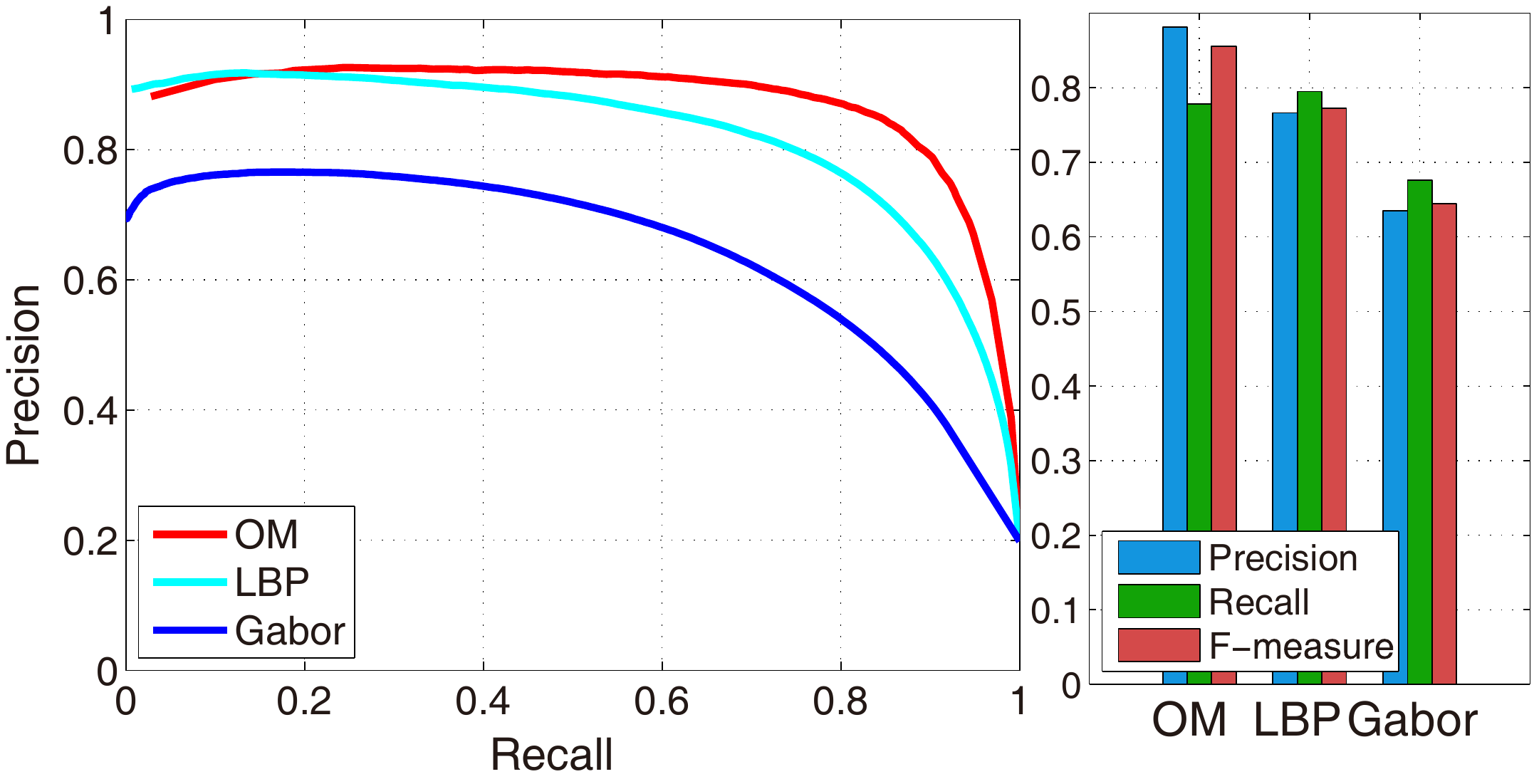}
\end{center}
\vspace{-15pt}
\caption
{\label{fig:grad}
Empirical study on two common structure features Gabor and LBP for replacing the proposed OM features in the PISA framework. Our OM descriptor performs better on the ASD dataset~\cite{ft-cvpr09}.}
\vspace{-10pt}
\end{figure}

We also justify the significance of the proposed energy-minimization framework (Eqn.~(\ref{equ:energyEqu})) by comparing with our conference framework (PISA-prev Framework)~\cite{pisa-cvpr13}. For fair comparison, we conduct the experiment with all other parameters fixed, i.e. they share the same normalized feature-based saliency measure $f(p)$, and the only difference is the framework. Fig.~\ref{fig:compFramework} demonstrates that our energy-minimization framework obtains higher precision when the recall belongs to [0, 0.2] on PR curves and achieves better MAE results. Thus, by modeling the appearance contrast based saliency measure and the neighborhood coherence constraint jointly, the proposed energy-minimization framework can highlight saliency objects more uniformly.

We have also explored other commonly used features Gabor~\cite{gabor-pami96} and LBP~\cite{lbp-pr09} to substitute OM for capturing structure information. For all the features, we choose their best results for comparison by tuning their quantizations. The dimensions for Gabor and LBP features are $72$ and $256$, respectively. The PR-curves of the experiments evaluated on the ASD dataset~\cite{ft-cvpr09} are shown in Fig.~\ref{fig:grad}. The OM descriptor outperforms the others. Meanwhile, under the proposed framework, our OM descriptor also shows higher computational efficiency than Gabor and LBP due to its low dimension.

\begin{figure}[htbp]
\begin{center}
\includegraphics[width = 1 \columnwidth]{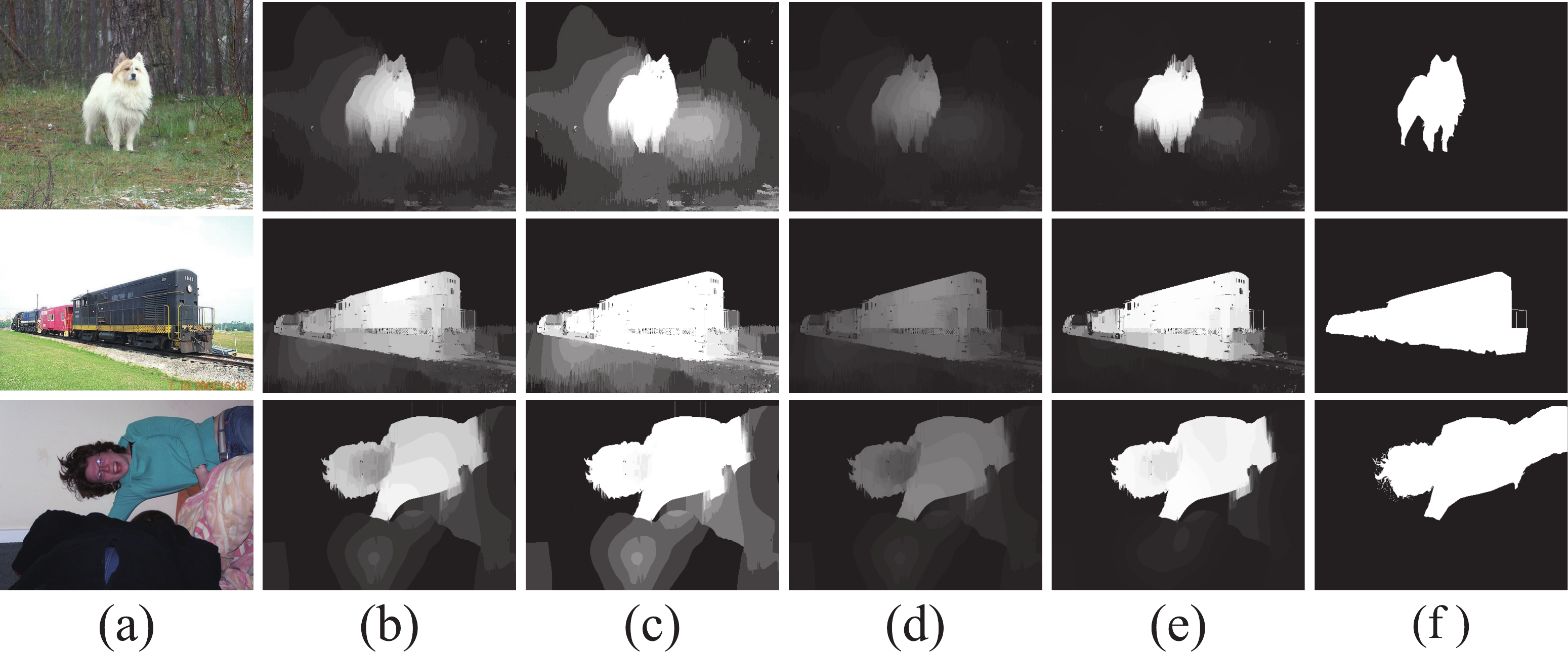}
\end{center}
\vspace{-15pt}
\caption
{\label{fig:norm_visual}
Visual results by different normalization methods. (a) Input image. (b) Max-min normalization. (c) Log-like normalization. (d) Exp-like normalization. (e) Sigmoid-like normalization. (f) Ground truth.}
\vspace{-14pt}
\end{figure}

\begin{figure}[!ht]
\centering
\includegraphics[width=0.95\columnwidth]{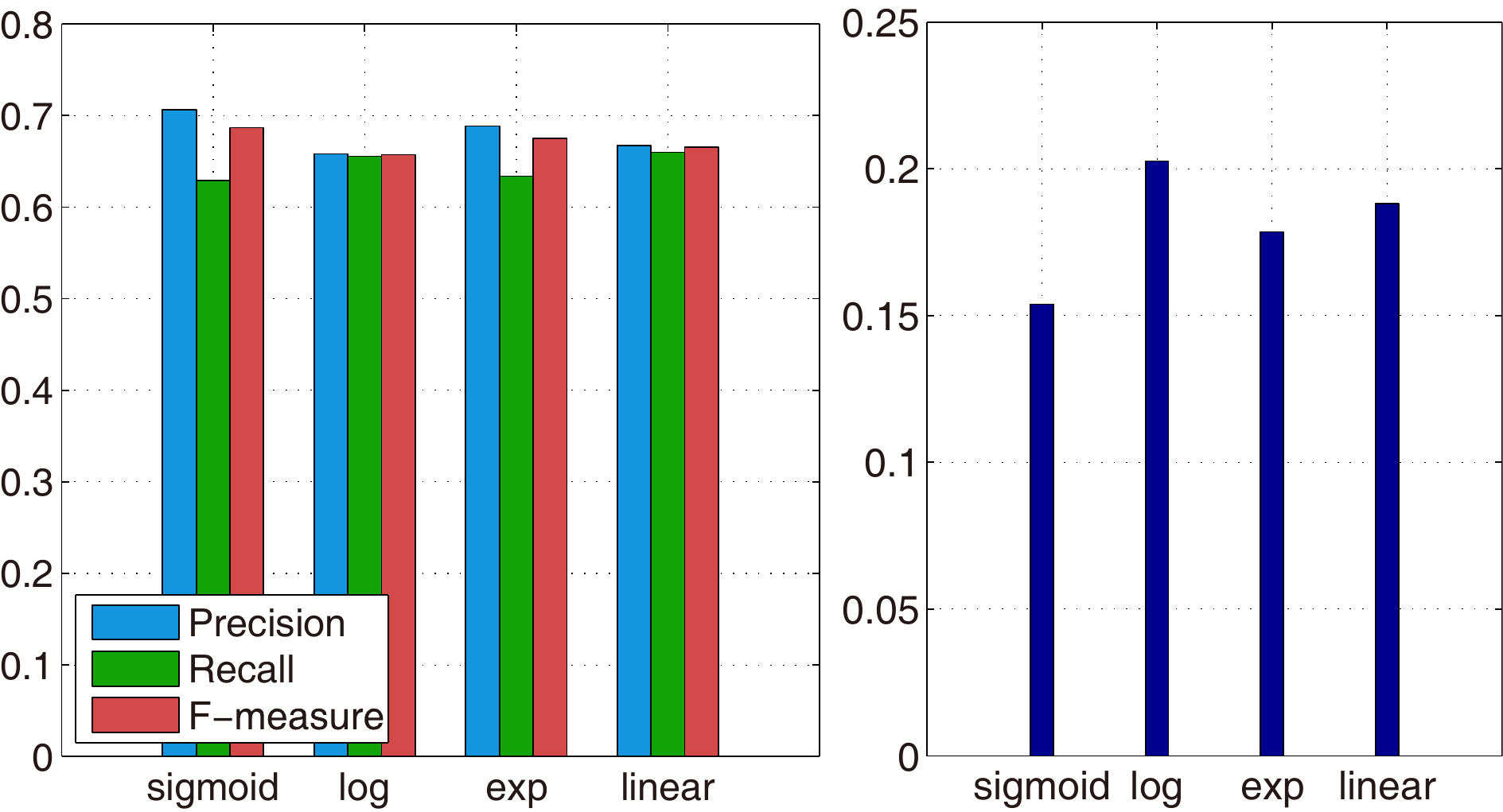}
\vspace{-10pt}
\caption{Quantitative results by different normalization methods on PASCAL-1500~\cite{LRR-bmvc13}. ``sigmoid'' denotes the sigmoid-like normalization while ``log'', ``exp'' and ``linear'' denote normalization of log-like, exp-like and max-min normalization, respectively.}\label{fig:normalization}
\vspace{-15pt}
\end{figure}

\begin{figure*} [htbp]
\begin{center}
\includegraphics[width =0.9\textwidth]{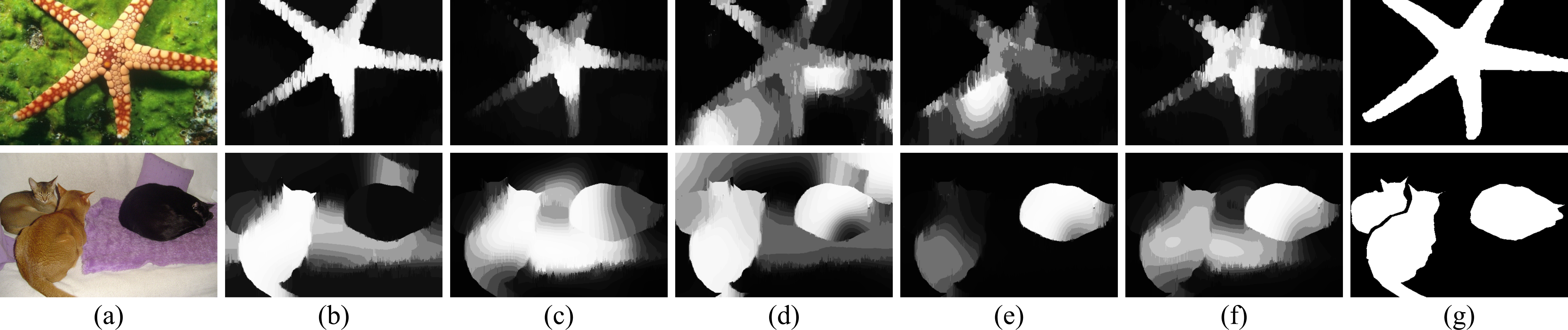}
\end{center}
\vspace{-12pt}
\caption
{\label{fig:limit}
An example case challenging for PISA. (a) Input image. (b) Color contrast measure. (c) Spatial prior-modulated color measure. (d) Structure contrast measure. (e) Spatial prior-modulated structure measure. (f) PISA result. (g) Ground truth.}
\vspace{-15pt}
\end{figure*}

In our proposed framework, the normalization step, which maps the feature-based saliency measure $\hat{f}(p)$ into discrete saliency level set \{0, ..., $\mathcal{L}-1$\}, has an impact on the final saliency maps. Fig.~\ref{fig:norm_visual} illustrates this impact of different normalizations, such as commonly used max-min (linear), log-like (nonlinear), and exp-like (nonlinear). Compared with the linear normalization, log-like increases the saliency levels of the whole pixels (Fig.~\ref{fig:norm_visual}(c)), while exp-like decreases all pixels' saliency levels (Fig.~\ref{fig:norm_visual}(d)). Sigmoid-like increases the number of high salient value pixels and reduces those of low salient value for its S shape (Fig.~\ref{fig:norm_visual}(e)). For exploring these normalization functions, we conduct the experiment on the PASCAL-1500 dataset~\cite{LRR-bmvc13} as it is the most challenging and the largest dataset with $F_{0.3}$ metric and MAE evaluation. Note that we discard the PR curves for that the change of normalization methods will not affect PR results as long as the mapping is one-to-one. Fig.~\ref{fig:normalization} demonstrates that a sigmoid-like function performs a little better in $F_{0.3}$ metric and much better in MAE evaluation than others. Thus we adopt a sigmoid-like normalization (Eqn.~(\ref{equ:sigmoid})) to produce better visual saliency maps. 

\subsection{Efficiency Analysis}
The experiments are carried out on a desktop with an Intel i7 3.4GHz CPU and 8GB RAM. The average runtime with ranking of our approaches (PISA and F-PISA) and competing methods on the ASD dataset~\cite{ft-cvpr09}, whose most images have a resolution of $300 \times 400$, are reported in Table~\ref{tab:time}. Though PISA is a little slow (rank 13, slightly faster than our conference version PISA-prev), our fast implementation F-PISA, significantly improves the efficiency (rank 6, $14$ times faster than PISA), while keeping comparable accuracy (better than the top five methods in the rank list, see Fig.~\ref{fig:PRF} and~\ref{fig:PRF2}). Specifically, for PISA: calculating the normalized feature-based saliency measure costs 310ms (about 50\%), minimizing the energy function costs 280ms (about 45\%), and others cost 30ms (about 5\%). For F-PISA: computing saliency costs 30ms (about 68\%), while subsampling and joint bilateral upsampling costs 12ms (about 27\%), and others cost 2ms (about 5\%).

\begin{table}\footnotesize
 \caption{Comparison of the average running time (seconds per image) on the ASD dataset~\cite{ft-cvpr09}.}
 \vspace{-8pt}
 \label{tab:time}
 \centering
 \begin{tabular}{|c|c|c|c|}
 \hline
 Approach             & Code Type     & Time   & Rank \\\hline
 LC \cite{lc-cvpr06}  & C++           & 0.004  & 1 \\\hline
 SR \cite{sr-cvpr07}  & C++           & 0.004  & 1 \\\hline
 HC~\cite{hcrc-cvpr11}& C++           & 0.009  & 3 \\\hline  
 FT \cite{ft-cvpr09}  & C++           & 0.010  & 4 \\\hline
 F-PISA-prev \cite{pisa-cvpr13} & C++ & 0.036  & 5\\\hline
 \textbf{F-PISA}     & C++            & 0.044  & 6 \\\hline
 RC~\cite{hcrc-cvpr11}& C++           & 0.090  & 7 \\\hline
 GC~\cite{gc-iccv13}  & C++           & 0.094  & 8 \\\hline
 GM \cite{GM-cvpr13}  & Matlab        & 0.111  & 9 \\\hline
 SF~\cite{sf-cvpr12}  & C++           & 0.136  & 10 \\\hline
 MC \cite{MC-iccv13}  & Matlab \& C++ & 0.150  & 11 \\\hline
 HS \cite{ca-cvpr10}  & C++           & 0.397  & 12 \\\hline
 \textbf{PISA}        & C++           & 0.620  & 13 \\\hline
 PISA-prev \cite{pisa-cvpr13} & C++   & 0.650  & 14 \\\hline	
 CB \cite{CB-bmvc11}  & Matlab \& C++ & 1.862  & 15 \\\hline
 DSR \cite{DSR-iccv13}& Matlab        & 3.536  & 16 \\\hline
 CA \cite{hs-cvpr13}  & Matlab        & 52.580 & 17 \\\hline
 \end{tabular}
 \vspace{-12pt}
 \end{table}

\subsection{Limitations}

In Fig.~\ref{fig:limit}, we present unsatisfying results generated by PISA. As our approach uses the spatial priors, it has problems when such priors are invalid. For example, if the saliency object occurs near the image boundary to quite extent, some regions of it can be suppressed (see Fig.~\ref{fig:limit}(first row)) due to the image boundary exclusion prior. If the center prior does not hold, the background regions located near the image center cannot be effectively suppressed in saliency evaluation (see Fig.~\ref{fig:limit}(second row)). By adjusting the relative contribution of these priors through tuning $\lambda$, we can alleviate their influences. Thus, the weakness of the proposed methods is: for any background regions that have been assigned high saliency values from either of the contrast cues after the modulation of the spatial priors, they remain salient in the final saliency map. This problem could be tackled by incorporating high-level knowledge to adjust the confidence of two measures in the formulation.

\section{Conclusion}
\label{sec:conclusion}
We have presented a generic and unified framework for pixelwise saliency detection by aggregating multiple image cues and priors, where the feature-based saliency confidence are jointly modeled with the neighborhood coherence constraint. Based on the saliency model, we employed the shape-adaptive cost-volume filtering technique to achieve fine-grained saliency value assignment while preserving edge-aware image details. We extensively evaluated our PISA on six public datasets by comparing with previous works. Experimental results demonstrated the advantages of our PISA in detection accuracy consistency and runtime efficiency. For future work, we plan to incorporate high-level knowledge and multilayer information, which could be beneficial to handle more challenging cases, and also investigate other kinds of saliency cues or priors to be embedded into the PISA framework.

\vspace{-40pt}
\begin{IEEEbiography}[{\includegraphics[width=1in,height=1.25in,clip,keepaspectratio]{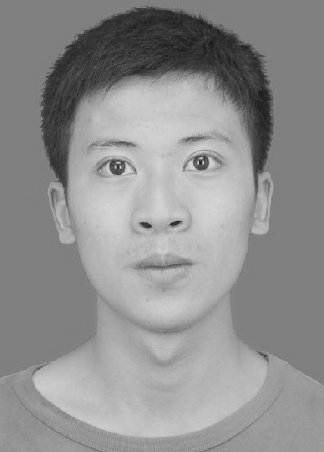}}]{Keze Wang}
received the BS degree in software engineering from Sun Yat-Sen University, Guangzhou, China, in 2012. He is currently pursuing the Ph.D. degree in computer science and technology at Sun Yat-Sen University, advised by Professor Liang Lin. His current research interests include computer vision and machine learning.
\end{IEEEbiography}

\vspace{-45pt}
\begin{IEEEbiography}[{\includegraphics[width=1in,height=1.25in,clip,keepaspectratio]{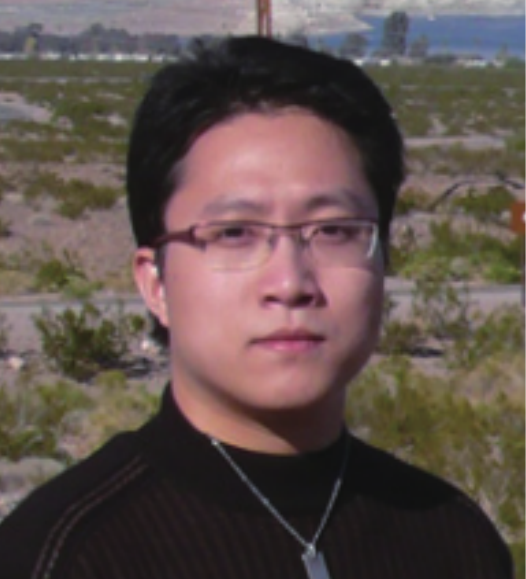}}]{Liang Lin} is a Professor with the School of Advanced Computing, Sun Yat-Sen University (SYSU), China. He received the B.S. and Ph.D. degrees from the Beijing Institute of Technology (BIT), Beijing, China, in 1999 and 2008, respectively. From 2006 to 2007, he was a joint Ph.D. student with the Department of Statistics, University of California, Los Angeles (UCLA). His Ph.D. dissertation was achieved the China National Excellent Ph.D. Thesis Award Nomination in 2010. He was a Post-Doctoral Research Fellow with the Center for Vision, Cognition, Learning, and Art of UCLA. His research focuses on new models, algorithms and systems for intelligent processing and understanding of visual data such as images and videos. He has published more than 70 papers in top tier academic journals and conferences. He was supported by several promotive programs or funds for his works, such as ``Program for New Century Excellent Talents'' of Ministry of Education (China) in 2012, and Guangdong NSFs for Distinguished Young Scholars in 2013. He received the Best Paper Runners-Up Award in ACM NPAR 2010, Google Faculty Award in 2012, and Best Student Paper Award in IEEE ICME 2014. He has served as an Associate Editor for Neurocomputing and The Visual Computer. 
\end{IEEEbiography}

\vspace{-45pt}
\begin{IEEEbiography}[{\includegraphics[width=1in,height=1.25in,clip,keepaspectratio]{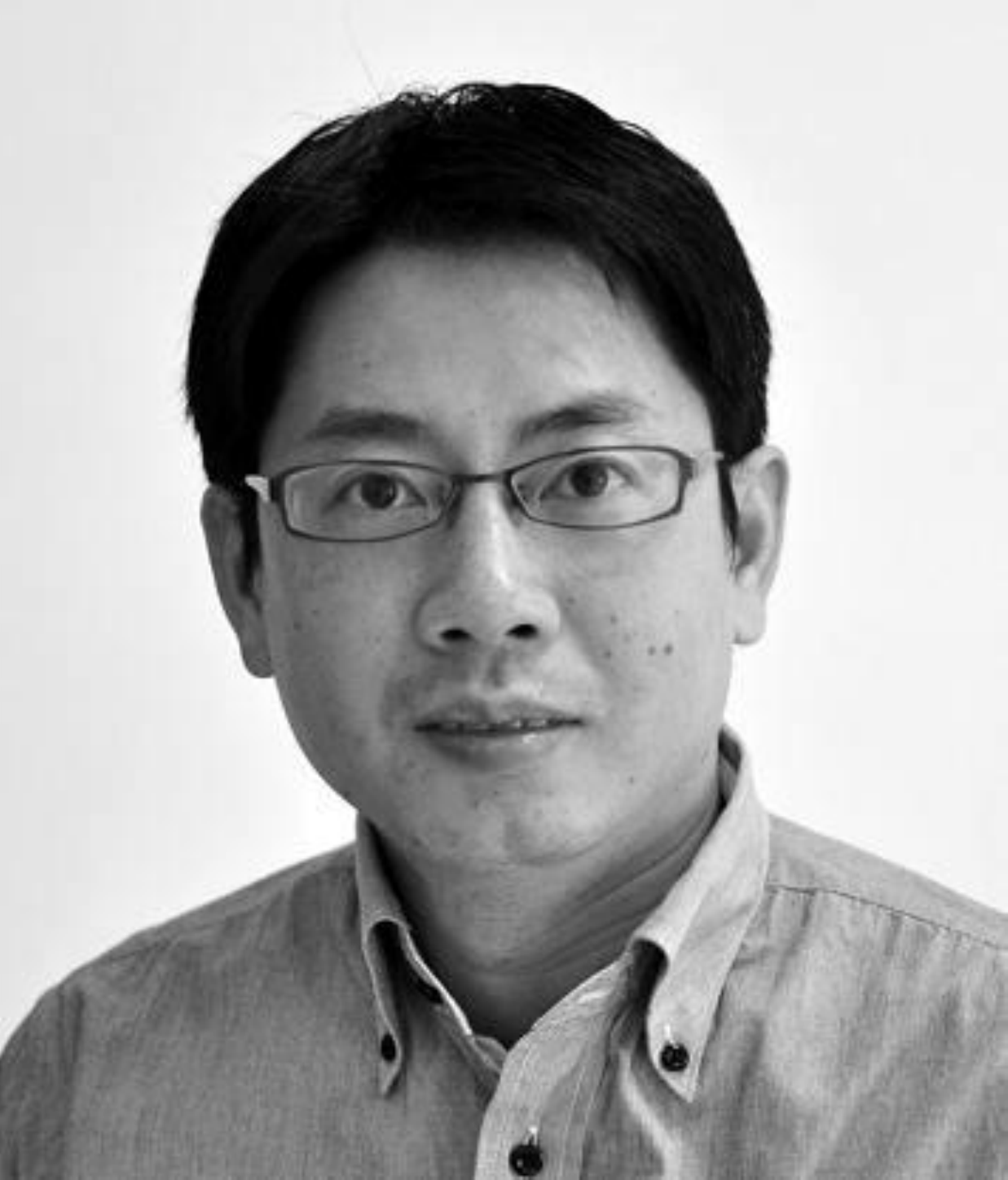}}]{Jiangbo Lu}(M'09) received the B.S. and M.S. degrees in electrical engineering from Zhejiang University, Hangzhou, China, in 2000 and 2003, respectively, and the Ph.D. degree in electrical engineering from Katholieke Universiteit Leuven, Leuven, Belgium, in 2009. From April 2003 to August 2004, he was with VIA-S3 Graphics, Shanghai, China, as a Graphics Processing Unit (GPU) Architecture Design Engineer. In 2002 and 2005, he conducted visiting research at Microsoft Research Asia, Beijing, China. Since October 2004, he has been with the Multimedia Group, Interuniversity Microelectronics Center, Leuven, Belgium, as a Ph.D. Researcher. Since September 2009, he has been with the Advanced Digital Sciences Center, Singapore, which is a joint research center between the University of Illinois at Urbana-Champaign, Urbana, and the Agency for Science, Technology and Research (A*STAR), Singapore, where he is leading a few research projects currently as a Senior Research Scientist. His research interests include computer vision, visual computing, image processing, video communication, interactive multimedia applications and systems, and efficient algorithms for various architectures.
\end{IEEEbiography}

\vspace{-45pt}
\begin{IEEEbiography}[{\includegraphics[width=1in,height=1.25in,clip,keepaspectratio]{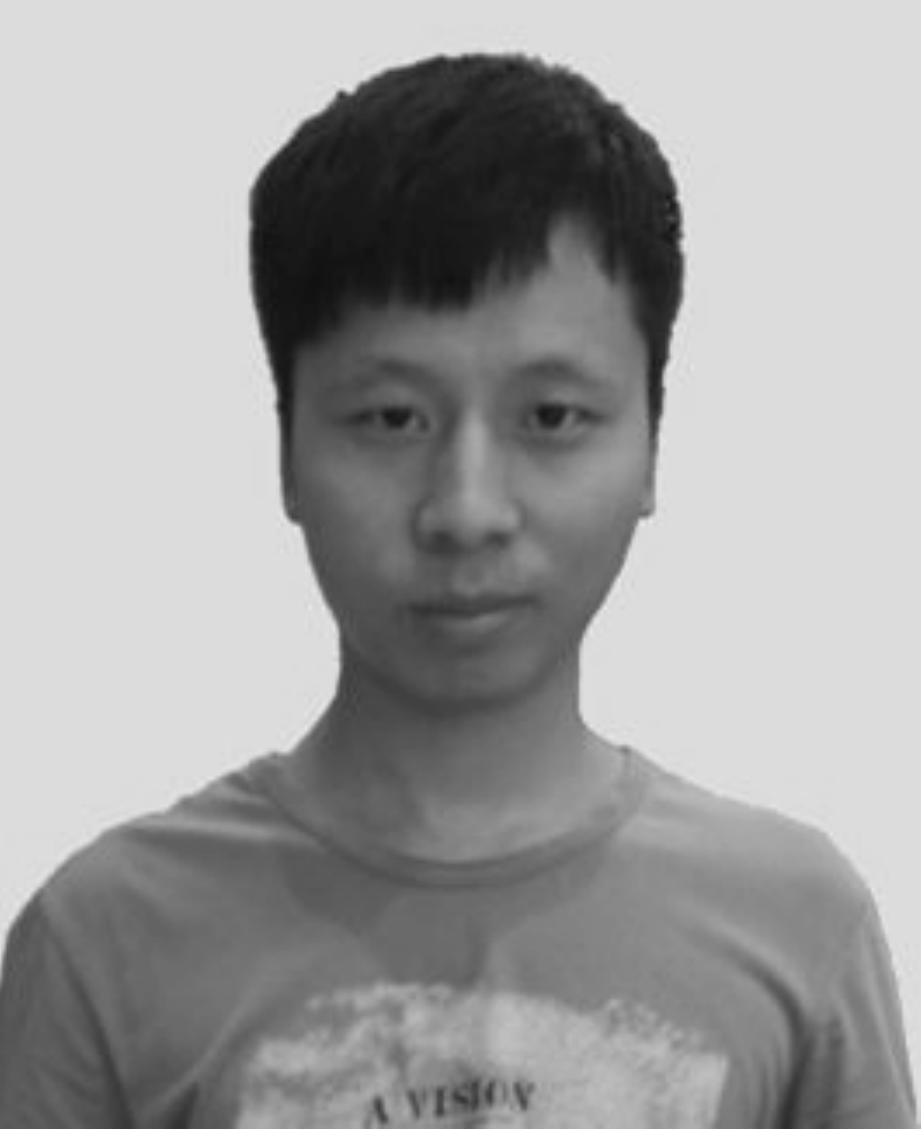}}]{Chenglong Li}
received the BS degree in applied mathematics in 2010, and the M.S. degree in computer science in 2013 from Anhui University, Hefei, China. He is currently pursuing the Ph.D. degree in computer science at Anhui University. His current research interests include computer vision, machine learning, and intelligent media technology.
\end{IEEEbiography}

\vspace{-45pt}
\begin{IEEEbiography}[{\includegraphics[width=1in,height=1.25in,clip,keepaspectratio]{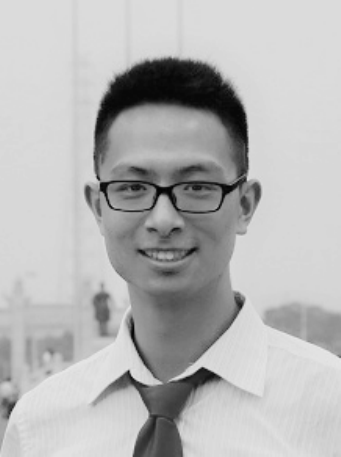}}]{Keyang Shi} 
received the BS and MS degrees in software engineering and computer science respectively, from Sun Yat-Sen University, Guangzhou, China, in 2011 and 2014. His research interests include computer vision, machine learning and cloud computing.
\end{IEEEbiography}
\end{document}